\PassOptionsToPackage{nopatch=footnote}{microtype}
\PassOptionsToPackage{table}{xcolor}
\documentclass[fleqn,11pt,a4paper]{wlscirep}
\usepackage{iftex}
\usepackage[mathlines]{lineno}
\usepackage[T1]{fontenc}
\ifPDFTeX
\usepackage[utf8]{inputenc}
\fi
\RequirePackage{XCharter}
\RequirePackage[xcharter,bigdelims,vvarbb]{newtxmath}
\DeclareMathAlphabet{\mathcal}{OMS}{cmsy}{m}{n}
\SetMathAlphabet{\mathcal}{bold}{OMS}{cmsy}{b}{n}
\RequirePackage[scaled=1.1]{zlmtt}
\DeclareFontFamily{OT1}{XCharter(0)}{}
\DeclareFontShape{OT1}{XCharter(0)}{m}{n}{<-> s * [1.0] XCharter-Roman-tlf-ot1}{}
\DeclareFontShape{OT1}{XCharter(0)}{b}{n}{<-> s * [1.0] XCharter-Bold-tlf-ot1}{}
\DeclareFontShape{OT1}{XCharter(0)}{m}{it}{<-> s * [1.0] XCharter-Italic-tlf-ot1}{}
\DeclareFontShape{OT1}{XCharter(0)}{b}{it}{<-> s * [1.0] XCharter-BoldItalic-tlf-ot1}{}
\DeclareFontShape{OT1}{XCharter(0)}{bx}{n}{<-> ssub * XCharter(0)/b/n}{}
\DeclareFontShape{OT1}{XCharter(0)}{bx}{it}{<-> ssub * XCharter(0)/b/it}{}

\newcommand{\footerfont}{\color{black}\normalfont\fontsize{10}{12}\selectfont}
\newcommand{\titlefont}{\color{black}\normalfont\bfseries\fontsize{14.5}{19}\selectfont}
\newcommand{\absfont}{\linespread{1.08}\fontsize{11}{13}\selectfont}
\newcommand{\headingfont}{\color{black}\bfseries\fontsize{13}{14}\selectfont}
\renewcommand\Affilfont{\normalfont\fontsize{11}{13}\selectfont\raggedright}
\usepackage{graphicx}
\usepackage{float}
\usepackage{tabularx}
\usepackage{array}
\usepackage{makecell}
\usepackage{multirow}
\newcommand{\choice}{\centering\arraybackslash\(\bigcirc\)}
\newcolumntype{N}{>{\centering\arraybackslash}m{0.045\linewidth}}
\newcolumntype{S}{>{\raggedright\arraybackslash}X}
\newcolumntype{R}{>{\centering\arraybackslash}m{0.070\linewidth}}
\definecolor{linklightblue}{HTML}{275D89}
\definecolor{frontiscyan}{HTML}{00A6D6}
\definecolor{frontisgray}{HTML}{4A4A4A}
\hypersetup{citecolor=linklightblue, linkcolor=linklightblue, urlcolor=linklightblue}
\titleformat{\section}
{\large\bfseries\headingfont}
{\thesection.}
{0.5em}
{#1}
[]
\titleformat{name=\section,numberless}
{\large\bfseries\headingfont}
{}
{0em}
{#1}
[]
\titleformat{\subsection}
{\color{black}\bfseries}
{\thesubsection.}
{0.5em}
{#1}
[]
\titleformat{\subsubsection}
{\color{frontisgray}\bfseries}
{\thesubsubsection.}
{0.5em}
{#1}
[]
\titlespacing*{\section}{0pc}{3ex plus4pt minus3pt}{5pt}
\titlespacing*{\subsection}{0pc}{2.5ex plus3pt minus2pt}{2pt}
\titlespacing*{\subsubsection}{0pc}{2ex plus2.5pt minus1.5pt}{2pt}
\titleformat{\paragraph}[runin]{\bfseries}{}{-0.5em}{#1}
\titlespacing*{\paragraph}{0pt}{0ex}{.33em}

\fancypagestyle{firststyle}{
  \fancyhead[L,R,C]{}
  \fancyfoot[L,R,C]{}

}
\makeatletter
\renewcommand\AB@authnote[1]{\textsuperscript{\fontencoding{T1}\normalfont\selectfont#1}}
\renewcommand\AB@affilnote[1]{\textsuperscript{\fontencoding{T1}\normalfont\selectfont#1}}
\renewcommand{\AB@affilsepx}{\protect\\\protect\Affilfont}
\newcommand{\affilnewline}{%
  \renewcommand{\AB@affilsepx}{\protect\\\protect\Affilfont}%
  \let\AB@affilsep\AB@affilsepx
}
\newcommand{\affilinline}{%
  \renewcommand{\AB@affilsepx}{\quad\protect\Affilfont}%
  \let\AB@affilsep\AB@affilsepx
}
\newcommand{\affilcorrespondinggap}{%
  \renewcommand{\AB@affilsepx}{\protect\\[0.55em]\protect\Affilfont}%
  \let\AB@affilsep\AB@affilsepx
}

\newcommand{\authornoaffil}[1]{%
  \begingroup
  \let\AB@authnote\@gobble
  \author[*]{#1}%
  \endgroup
}
\renewcommand{\@maketitle}{%
{%
\thispagestyle{firststyle}%
\vskip-14pt%
{\raggedright\titlefont \@title\par}%
\vskip0pt
{\raggedright\@author\par}%
\vskip22pt%
{%
\noindent
{\parbox{\dimexpr\linewidth-2\fboxsep\relax}{\color{color1}\large\textbf{Abstract}}}
}%
\vskip0pt
{%
\noindent
\colorbox{color2}{%
\parbox{\dimexpr\linewidth-2\fboxsep\relax}{%
\absfont\theabstract
}%
}%
}%
}%
}
\makeatother
\title{Towards AI-Assisted Clinical Trial Matching: Practical Considerations, Multicenter Evaluation, and Real-World Deployment}

\author[1]{Yin Fang}
\author[1]{Qiao Jin}
\author[1]{Shubo Tian}
\author[1]{Lauren He}
\author[1]{Maya Geer}
\author[2]{Noor Naffakh}
\author[2]{Ryan Huu-Tuan Nguyen}
\author[3]{Zifeng Wang}
\author[3]{Jimeng Sun}
\author[4]{Charalampos S. Floudas}
\author[4]{James L. Gulley}
\author[5]{Kamilia Moalem}
\author[5]{Catarina Martins Maia}
\author[6]{Amanda Nottke}
\author[6,9]{Juan W. Valle}
\author[6]{Melinda Bachini}
\author[6]{Lourdes Rocha-Nussbaum}
\author[6]{Kari Ramage}
\author[7]{Nikita Curry}
\author[7]{Megan Barnes}
\author[7]{Mandy Mansaray}
\author[8]{Darlene Gabeau}
\author[8]{Craig E. Grossman}
\author[8]{Heath Skinner}
\author[8]{Michael Burczynski}
\authornoaffil{NIH-TrialBench Consortium}
\author[1,*]{Zhiyong Lu}
\affil[1]{Division of Intramural Research, National Library of Medicine, National Institutes of Health, Bethesda, MD, USA.}
\affil[2]{UI Health, University of Illinois Chicago. Chicago, IL, USA.}
\affil[3]{University of Illinois Urbana-Champaign, Urbana, IL, USA.}
\affil[4]{Center for Immuno-Oncology, Center for Cancer Research, National Cancer Institute, National Institutes of Health, Bethesda, MD, USA.}
\affilnewline
\affil[5]{Medical Oncology Service, Center for Cancer Research, National Cancer Institute, National Institutes of Health, Bethesda, MD, USA}
\affil[6]{The Cholangiocarcinoma Foundation, Herriman, UT, USA.}
\affil[7]{Office of Patient Recruitment, NIH Clinical Center, Bethesda, MD, USA.}
\affil[8]{UPMC Hillman Cancer Center, University of Pittsburgh, Pittsburgh, PA, USA.}
\affil[9]{Division of Cancer Sciences, University of Manchester, Manchester, UK.}
\affilcorrespondinggap
\affil[*]{Corresponding author: zhiyong.lu@nih.gov}

\begin{abstract}
Clinical trials are essential for evidence-based medicine and drug development, but many fail because of insufficient patient enrollment. Although there is growing interest in using AI to support patient recruitment, existing systems largely perform eligibility assessment alone and have rarely been evaluated in real-world workflows. Here we present TrialGPT 2.0, an AI-assisted clinical trial recommendation system improved for responsible real-world deployment. Rather than asking only whether a patient may qualify, the system assesses which trials warrant further consideration given the patient's current clinical needs and local workflow priorities, and provides structured, inspectable explanations for expert review. Importantly, we evaluated TrialGPT 2.0 retrospectively and prospectively across government research, patient-advocacy, academic medical center, and community-based radiation-oncology settings.
In retrospective multicenter cohorts comprising 288 cases, TrialGPT 2.0 retrieved at least one clinician-recommended trial in its top 10 recommendations for approximately 91\% of cases while reducing clinician screening time by 55.0\%.
In a six-month prospective evaluation embedded in an active precision oncology tumor board, TrialGPT 2.0 contributed additional trial opportunities missed by the routine workflow, accounting for 69\% of final clinician-selected trial recommendations and increasing the number of patient cases with at least one final trial recommendation by 90.9\%.
To support scientific reproducibility, we also developed NIH-TrialBench, a clinician-authored benchmark comprising 126 diverse synthetic patient vignettes and matching scenarios from 11 NIH Institutes and Centers. 
Together, these findings support the value of AI to assist clinical trial matching by improving clinician efficiency and identifying frequently overlooked trial opportunities, ultimately helping to expand and accelerate accrual to cancer trials. TrialGPT 2.0 has been deployed in real-world trial-review workflows.

\end{abstract}
\begin{document}

\raggedbottom
\maketitle
%
%
\thispagestyle{firststyle}
\vspace{3pt}

\section*{Introduction}

Clinical trials provide evidence that guides medical practice and remain the gold standard for drug development. However, bringing a new drug to market is resource-intensive, often requiring over a decade and more than a billion dollars, with clinical trials accounting for over half of these expenditures~\cite{dimasi2016innovation,sertkaya2024costs}.
Patient-trial matching is a central operational step in recruitment, particularly in oncology, where it requires interpretation of complex and rapidly evolving clinical and molecular eligibility criteria, integration of heterogeneous patient information for each candidate trial, and continuous review of the changing portfolio of trials available at a given site~\cite{unger2019systematic}. 
Today, this step remains largely manual, making it slow and error-prone~\cite{penberthy2012effort}. Poor patient recruitment is a leading driver of research waste in randomized controlled trials (RCTs), accounting for 36.7\% of premature discontinuations, which affect 30.1\% of general RCTs~\cite{speich2025nonregistration}. The challenge is especially acute in oncology: 22.7\% of interventional cancer trials terminate early, with accrual problems responsible for 34.5\% of those terminations~\cite{hauck2021trial,zhang2023early}, and low accrual accounts for 57.0\% of failed Phase II/III cancer trials~\cite{schroen2012achieving,peterson2023growth}. 

To address this bottleneck, various artificial intelligence (AI) methods have been explored to support patient recruitment through trial search~\cite{jin2024matching,rybinski2024learning,chan2025recommending,abdallah2026trialmatchai}, cohort identification~\cite{park2024criteria2query,klein2022matchminer} and eligibility screening~\cite{zhang2020deepenroll,beck2020artificial,alexander2020evaluation,nievas2024distilling,gupta2024prism,datta2024autocriteria,beattie2024utilizing,bornet2025analysis,wornow2025zero,chen2025enhancing}. For example, we previously developed TrialGPT, a first-of-its-kind large language model (LLM)-based framework for zero-shot patient-to-trial matching (i.e., without task-specific training) that combines retrieval, criterion-level eligibility assessment, and trial-level ranking into a single system~\cite{jin2024matching}.

Despite this progress, existing AI-assisted patient-trial matching approaches share two major limitations that constrain their clinical value. First, they remain focused on trial eligibility rather than clinical recommendation. These are distinct questions: eligibility establishes whether a patient could plausibly enroll, whereas recommendation reflects whether the trial is appropriate to pursue given the patient's current clinical needs. This distinction is especially important in oncology, where a patient may be eligible for many trials but warrant recommendation for only a few (e.g., a clinician might not recommend an eligible trial because it is purely diagnostic, offers little expected benefit relative to standard care, or contradicts the patient's overall treatment plan). Therefore, eligibility is necessary but not sufficient for recommendation.
Second, prior evaluations have relied largely on technical validation using synthetic or eligibility-oriented benchmarks and have provided limited evidence from clinician-adjudicated or prospective review workflows with real patient data, leaving it unclear whether AI-identified trials contribute useful options to real-world trial-selection decisions~\cite{beck2020artificial,alexander2020evaluation}. Together, these gaps point to an urgent and unmet need in medical AI research: to evaluate patient-trial matching as a clinical recommendation task aligned with routine clinical workflows, rather than on benchmark performance alone~\cite{kann2021artificial,kehl2024identifying,showus2026}.

To this end, we present TrialGPT 2.0, an enhanced TrialGPT~\cite{jin2024matching} system designed to assist clinicians and trial specialists in real-world patient-trial matching. Unlike its predecessor, TrialGPT 2.0 is designed with greater configurability to support the trial-review needs of different recruitment settings and workflows. Each setting uses a predefined trial list relevant to its workflow (e.g., an institutional or program-specific portfolio) and a matching policy that specifies how potentially eligible trials should be prioritized (e.g., by favoring stronger disease or biomarker matches or therapeutic rather than diagnostic or registry trials). Given a patient, TrialGPT 2.0 generates a ranked list of candidate trials based on overall patient-trial fit. Each result includes a structured assessment separating eligibility-supporting evidence, potential incompatibilities, and missing information, alongside a rationale to support transparent clinical review.

More importantly, we evaluated TrialGPT 2.0 across complementary retrospective and prospective cohorts.
First, we performed a multicenter retrospective evaluation of 1{,}340 heterogeneous patient-trial pairs from 288 de-identified clinical-note cases spanning government research, patient-advocacy, academic cancer-center, and community-based radiation-oncology settings. These cases covered referral, intake, consultation, and tumor-board workflows and varied in patient characteristics, documentation length, and local trial search space. Second, we prospectively assessed TrialGPT 2.0 in a Precision Oncology Tumor Board (POTB) trial-review task~\cite{gueguen2025prospective}. From February through July 2026, TrialGPT 2.0 outputs were reviewed during routine clinical practice across 27 cases and 339 patient-trial pairs to assess whether the system contributed final trial options beyond the conventional human workflow. These evaluations directly address the two limitations identified above: they assess recommendation quality beyond eligibility through domain-expert clinical review, and they test whether AI-identified options are selected into final recommendations in a real-world workflow.

Additionally, to support reproducible evaluation despite restrictions on sharing real clinical data, we developed NIH-TrialBench, a clinician-authored benchmark comprising 126 synthetic vignettes contributed by investigators from 11 NIH Institutes and Centers. We evaluated TrialGPT 2.0 on both NIH-TrialBench and established public patient-trial matching benchmarks~\cite{koopman2016test,soboroff2021overview,roberts2022overview}.

Finally, we implemented and deployed these capabilities in a web-based interface for routine trial review in real-world clinical workflows.
Together, these results support AI-assisted matching as a practical means of helping clinical teams identify clinically appropriate trial options more efficiently, an important step toward broadening and accelerating clinical trial recruitment.

\section*{Results}

\subsection*{TrialGPT 2.0 provides configurable and explainable clinical trial recommendations}

\begin{figure}[!t]
  \centering
  \includegraphics[width=\textwidth]{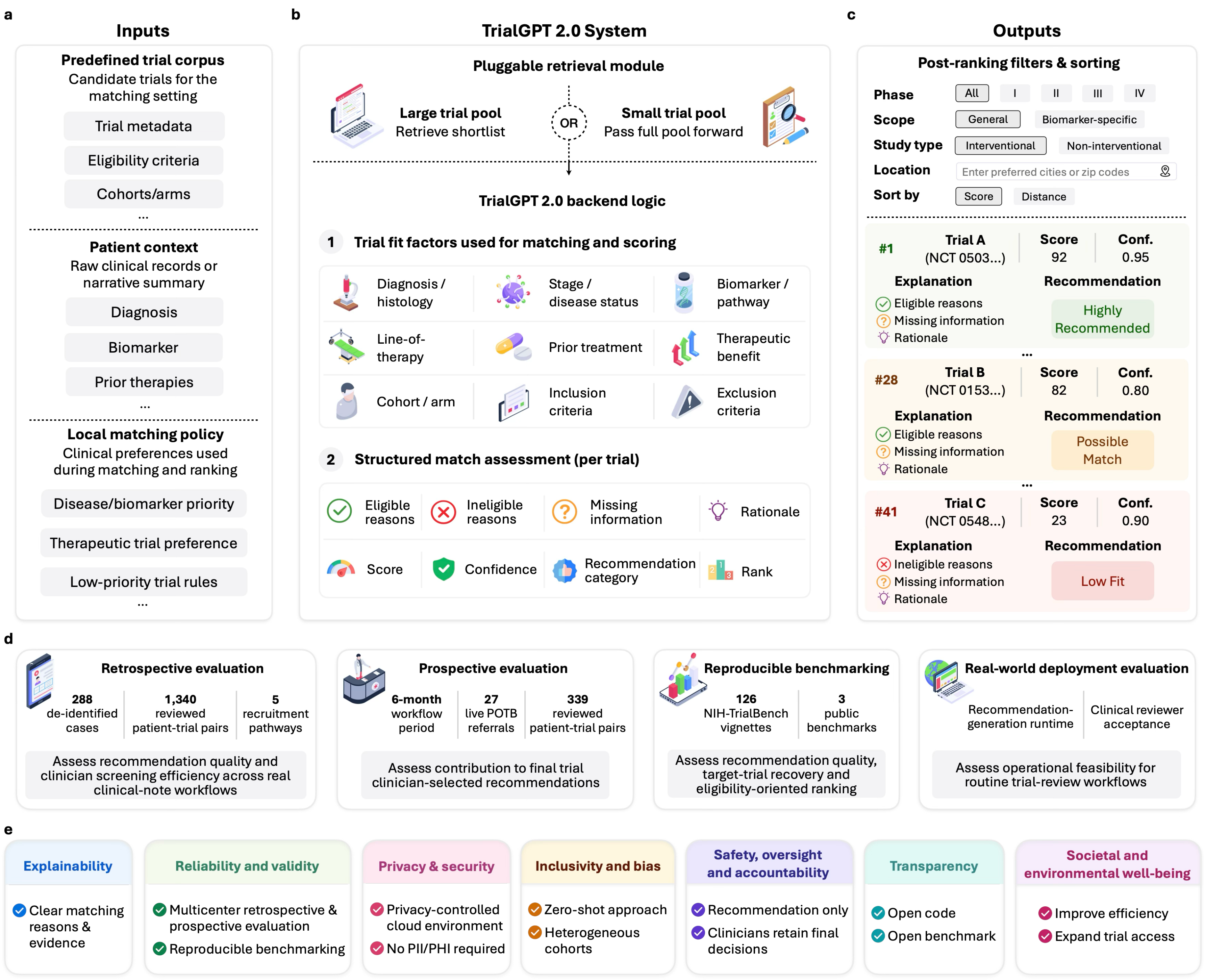}
  \caption{\textbf{Overview of the TrialGPT 2.0 system, evaluation framework, and responsible AI considerations.}
  \textbf{a,} Inputs for patient-trial matching, including the local trial corpus, patient context and center-specific matching policies.
  \textbf{b,} TrialGPT 2.0 system architecture. A pluggable retrieval module supports trial corpora of different sizes, and the backend evaluates trial-fit factors to generate structured trial-level assessments, including eligibility evidence, missing information, rationale, fit score, confidence, recommendation and rank.
  \textbf{c,} Ranked output interface with post-ranking filters, sorting options, trial-level explanations, scores, confidence estimates and recommendation categories.
  \textbf{d,} Evaluation framework. TrialGPT 2.0 is evaluated through retrospective clinician-adjudicated review, prospective workflow evaluation, reproducible benchmarking, and real-world deployment evaluation. 
  \textbf{e,} TrialGPT 2.0 design, evaluation, and governance measures addressing seven ethical elements relevant to responsible AI-assisted clinical trial matching~\cite{rentzepis2026use}.
  }
  \label{fig:overview}
\end{figure}

TrialGPT 2.0 is an AI-powered clinical trial matching system that assists clinicians in identifying and recommending clinical trials for individual patients. Here, recommendation refers not only to apparent eligibility but also to whether a trial warrants further consideration given the patient's current clinical needs. 
For input, the system takes patient information, a predefined list of available trials relevant to the recruitment setting (e.g., an institutional portfolio or a list identified through a disease-focused or geographic query), and a local matching policy that reflects the clinical preferences (e.g., disease- or biomarker-specific priorities, or a preference for therapeutic over diagnostic or registry trials) for prioritizing eligible trials (Fig.~\ref{fig:overview}a). For output, TrialGPT 2.0 generates a ranked list of trial recommendations for clinical review (Fig.~\ref{fig:overview}c).

To accommodate varying numbers of applicable trials at different sites and workflows, TrialGPT 2.0 applies a pluggable retrieval component before detailed matching assessment. For local lists exceeding 1{,}500 trials, the retrieval module is automatically activated to generate a candidate shortlist for improved efficiency, using the previously described hybrid-fusion strategy of TrialGPT that combines lexical and semantic retrieval~\cite{jin2024matching}; otherwise, the full list of trials is passed directly to the backend. 
The backend then generates an assessment for each candidate trial by comparing the patient with the trial's own inclusion and exclusion criteria and evaluating relevant trial-fit factors defined by the matching policy for the recruitment setting. These factors may include diagnosis or histology, disease status, biomarker or pathway match, line of therapy, prior treatment, and cohort or arm applicability. Each matching policy reflects the clinical priorities and information available in the corresponding recruitment setting by specifying which factors are explicitly assessed and which are considered only when relevant information is available (\hyperref[suppnote:factor]{Supplementary Note 1}).
Subsequently, for each candidate trial, TrialGPT 2.0 returns a recommendation category (Highly Recommended, Possible Match, or Low Fit), a rank, a fit score, a confidence estimate, and a structured explanation separating eligible versus ineligible reasons, missing information, and rationale, making both the recommendation and its supporting evidence inspectable (Fig.~\ref{fig:overview}b). To facilitate review, we implemented a web-based interface that presents ranked trial cards with these structured explanations and supports customizable post-ranking filtering and sorting aligned with local workflow constraints and referral priorities (Supplementary Fig.~\ref{fig:interface}).

\begin{figure}[p]
  \centering
  \includegraphics[width=0.98\textwidth]{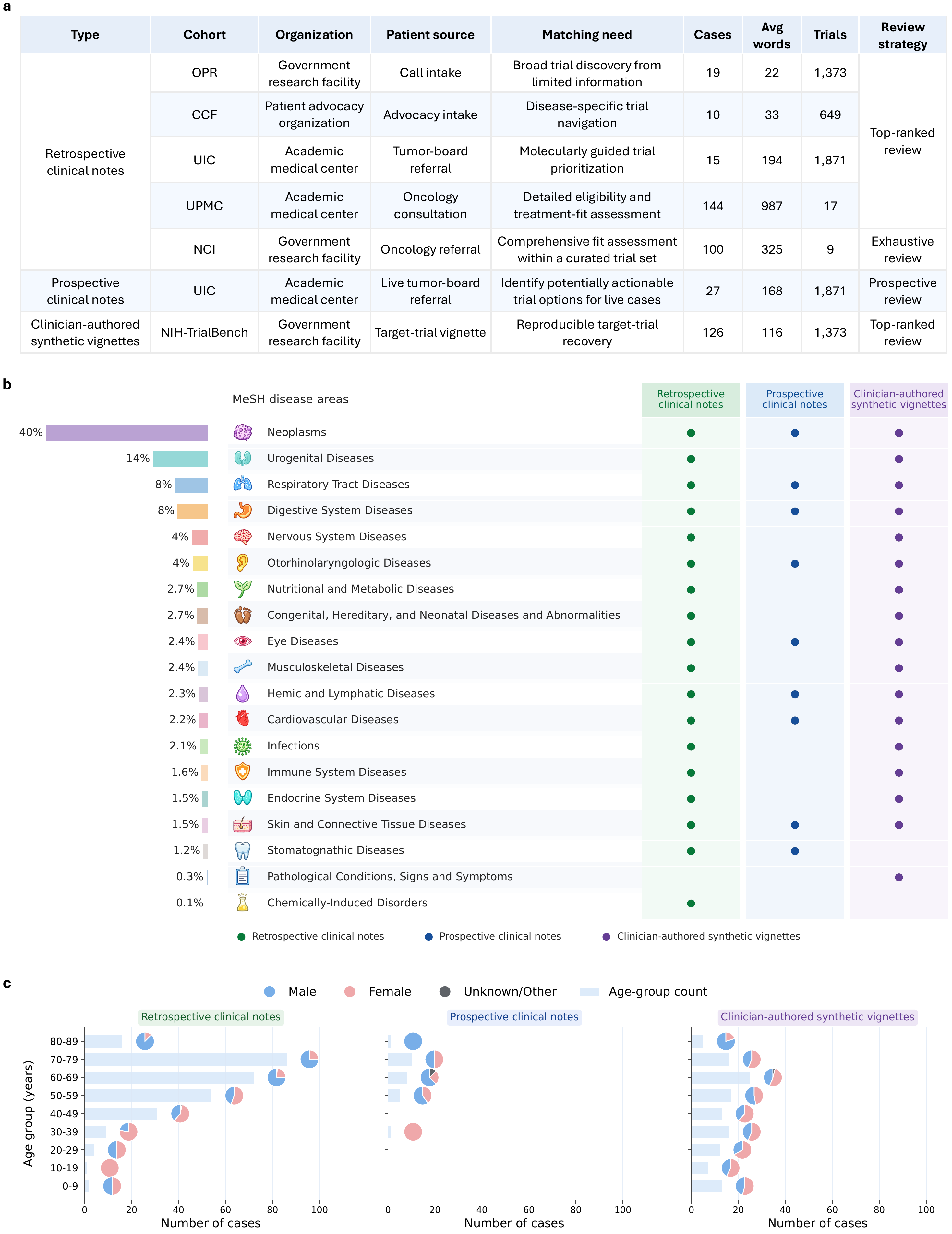}
  \caption{\textbf{Characteristics of study cohorts.}
    \textbf{a,} Summary of study cohorts. ``Top-ranked review'' indicates clinician labeling of up to 10 top-ranked TrialGPT 2.0 Highly Recommended trials per case. ``Exhaustive review'' indicates clinician labeling of all trials in the search space. ``Prospective review'' indicates workflow-based review in the prospective POTB setting.
    \textbf{b,} Disease-area coverage across cohorts. Patient contexts are mapped to first-level MeSH disease categories, with individual cases allowed to span multiple disease areas. Horizontal bars show the percentage of disease-area assignments in each MeSH category, and dots indicate whether each category was represented in each cohort type.
    \textbf{c,} Age and sex distributions across study cohorts. Horizontal bars show age-group counts, and overlaid pie markers indicate sex composition within each age group.
  }
  \label{fig:cohorts}
\end{figure}

We evaluated TrialGPT 2.0 through four complementary components (Fig.~\ref{fig:overview}d), with the clinical cohorts spanning government research, patient-advocacy, academic cancer-center, and community-based radiation-oncology settings (Fig.~\ref{fig:cohorts}a). Across the study cohorts, the local trial list ranged from 9 to 1{,}871 candidate trials, scoped by disease or biomarker focus, geographic region, or institutional availability (Fig.~\ref{fig:cohorts}a).

First, we performed a retrospective, clinician-adjudicated review of recommendation quality across de-identified clinical-note workflows. This evaluation comprised 288 de-identified cases from five recruitment pathways: patient call intake notes from the NIH Office of Patient Recruitment (OPR); oncology patient advocacy intake notes from the Cholangiocarcinoma Foundation (CCF); tumor-board referrals from the University of Illinois Cancer Center (UIC); radiation-oncology consultation notes from a community cancer program within the University of Pittsburgh Medical Center (UPMC) Hillman Cancer Center; and oncology referral notes from the National Cancer Institute (NCI). These pathways represented distinct matching needs: broad referral triage; disease- and biomarker-specific navigation for cholangiocarcinoma and biliary tract cancer; molecularly guided trial prioritization in precision oncology; radiation-oncology treatment-fit assessment in a community oncology setting; and comprehensive referral screening within a focused institutional oncology portfolio. The patient documents differed substantially in format and detail, ranging from short OPR and CCF intake or advocacy summaries and concise UIC molecular tumor-board referrals to curated NCI referral summaries and longer UPMC narrative or document-derived clinical notes.
Each site defined its own candidate trial list using a workflow-appropriate scoping strategy. CCF and UIC derived their lists from ClinicalTrials.gov~\cite{clinicaltrials_gov} using disease- and biomarker-specific and location-based queries, respectively; OPR used the NIH Clinical Center protocol inventory~\cite{nih_search_studies}, NCI used a focused institutional oncology set, and UPMC used a manually curated radiation-oncology portfolio. Matching was performed independently for each setting against its predefined list, reflecting how trial review is scoped in practice, where clinical teams generally consider a relevant institutional, geographic, disease-focused, or program-specific trial set rather than the full registry. Accordingly, most search spaces contained fewer than 1{,}500 candidate trials (Fig.~\ref{fig:cohorts}a).

Second, we prospectively evaluated whether TrialGPT 2.0-identified options contributed to final recommendations in the University of Illinois Cancer Center Precision Oncology Tumor Board (UIC POTB), a routine multidisciplinary oncology workflow for reviewing referred cases and identifying clinical trial options. The evaluation included 27 de-identified cases. For each case, TrialGPT 2.0 and clinical reviewers identified candidate trials in parallel from the same patient context. The two candidate-trial lists were then reviewed together, and the board selected the final trial recommendations. TrialGPT 2.0 used the UIC trial-corpus snapshot available at the time of each case review, which was refreshed weekly to reflect changes in trial availability, recruitment status, and cohort or arm availability (Fig.~\ref{fig:cohorts}a).

Third, we conducted a technical validation on existing eligibility-focused public patient-trial matching benchmarks and on NIH-TrialBench, a new benchmark we developed comprising 126 synthetic vignettes spanning oncology and other disease areas. The vignettes were created by 24 NIH trial investigators across 11 NIH Institutes and Centers to reflect real-world considerations in trial enrollment and screening. These vignettes were evaluated against a predefined search space of 1{,}373 candidate trials from the NIH Clinical Center's \textit{Search the Studies} database~\cite{nih_search_studies}, and clinicians reviewed and labeled the top-ranked TrialGPT 2.0 recommendations for each vignette. Unlike the data in the first two evaluations, NIH-TrialBench is released to the research community to support open science and reproducibility (Fig.~\ref{fig:cohorts}a).

Fourth, we evaluated real-world deployment through a pilot user survey of clinical reviewer acceptance and interface runtime testing (Fig.~\ref{fig:overview}d).

Together, these four components were designed to provide complementary coverage of evaluation settings, disease areas, patient demographics, and documentation patterns. Neoplasms were the largest disease-area category, reflecting the predominant oncology focus of the clinical workflows, while the combined evaluation also spanned a broad range of Medical Subject Headings (MeSH)~\cite{lowe1994understanding} categories. Retrospective clinical notes and clinician-authored synthetic vignettes showed the broadest disease-area coverage, whereas prospective cases were concentrated in the precision oncology setting (Fig.~\ref{fig:cohorts}b). Age and sex distributions also differed across cohort types (Fig.~\ref{fig:cohorts}c).

\subsubsection*{TrialGPT 2.0 incorporates responsible AI practices and safeguards}

Responsible AI-assisted clinical trial matching requires attention not only to model performance but also to ethical considerations arising from system design, evaluation, deployment, and governance. TrialGPT 2.0 incorporates measures that address all seven ethical elements previously identified for AI-enabled clinical trial recruitment (Fig.~\ref{fig:overview}e)~\cite{rentzepis2026use}.

Explainability is addressed through structured trial-level outputs that provide clear matching rationales and supporting evidence. Reliability and validity are addressed through multicenter retrospective evaluation, prospective workflow evaluation, and reproducible benchmarking. Privacy and security are addressed through deployment in a HIPAA-compliant cloud environment and data-minimization practices. For the web interface, only the minimum patient information needed for matching is used; personally identifiable information (PII) and protected health information (PHI) are not required, and user inputs are retained for no longer than 24 hours. The study evaluations used de-identified or synthetic patient contexts under applicable institutional oversight. Inclusivity and bias are addressed through a zero-shot approach requiring no task-specific training data or examples, together with evaluation across heterogeneous cohorts. Safety, oversight, and accountability are addressed through use of TrialGPT 2.0 as a recommendation-support tool only, with clinicians retaining responsibility for final trial recommendations, referral, screening, and enrollment decisions. Transparency is addressed through public release of NIH-TrialBench and the code and configurations used for reproducible benchmarking. Societal and environmental well-being are addressed through a design intended to augment rather than replace clinicians, together with evaluation of clinician screening efficiency, additional trial opportunities, and computational inference burden.

\subsection*{TrialGPT 2.0 returns clinician-recommended trials and improves screening efficiency in multicenter retrospective evaluations}

\begin{figure}[p]
  \centering
  \includegraphics[width=\textwidth]{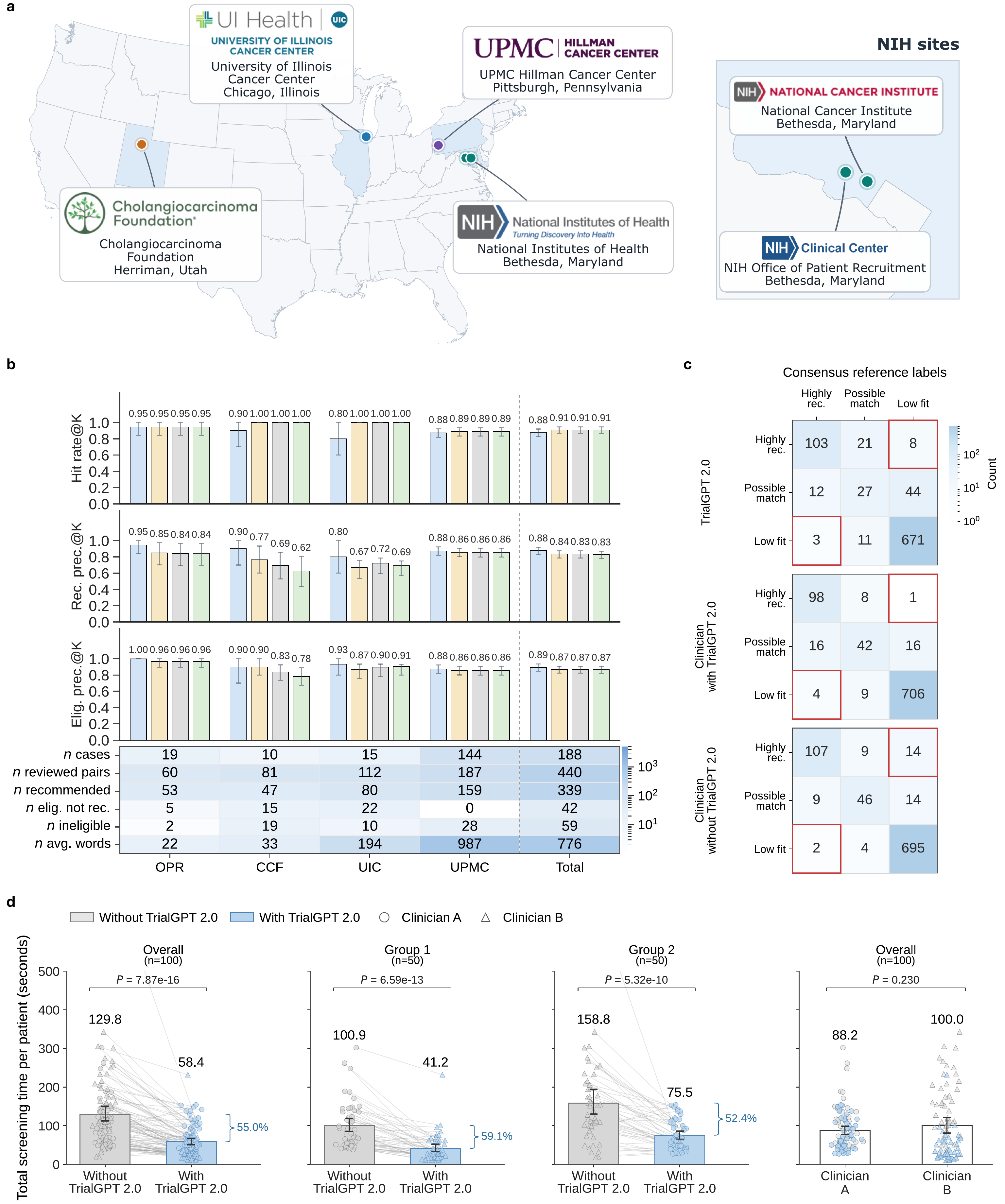}
  \caption{\textbf{Retrospective evaluation of TrialGPT 2.0 recommendations and clinician screening time.}
    \textbf{a,} Geographic locations of the participating retrospective cohort sites across the United States.
    \textbf{b,} Recommendation quality performance on retrospective clinical-note cohorts from OPR, CCF, UIC, and UPMC. Bars show hit rate@$K$, recommended precision@$K$, and eligible precision@$K$ for $K = 1, 3, 5$ and $10$, with error bars showing 95\% confidence intervals of the mean computed by percentile bootstrap (10{,}000 resamples).
    \textbf{c,} Agreement with consensus reference labels in the NCI exhaustive review setting. Red boxes mark severe reversals between Highly Recommended and Low Fit.
    \textbf{d,} Total trial-level screening decision time per case in the NCI counterbalanced experiment. Points represent cases, lines connect paired assisted and unassisted reviews, and brackets show percentage reductions. Bars show means and error bars show 95\% bootstrap confidence intervals (10{,}000 resamples). Statistical comparisons used two-sided Wilcoxon signed-rank tests for paired reviews and a two-sided Mann-Whitney \(U\) test between clinicians. All 100 cases were analyzed; one outlier falls outside the displayed range.
  }
  \label{fig:retrospective}
\end{figure}

We evaluated TrialGPT 2.0 in retrospective clinical-note cohorts from five recruitment settings across the United States (Fig.~\ref{fig:retrospective}a) to assess whether its recommendations aligned with clinician judgment and whether TrialGPT 2.0 assistance reduced clinician screening time.

We first assessed top-ranked recommendations in the OPR, CCF, UIC, and UPMC cohorts, spanning broad NIH referral triage, cholangiocarcinoma trial navigation, precision-oncology tumor-board review, and community-based radiation-oncology consultation. Clinicians reviewed up to ten top-ranked trials categorized by TrialGPT 2.0 as Highly Recommended (or all such trials if the system identified fewer than ten), and labeled each patient-trial pair as Recommended, Eligible but Not Recommended, or Ineligible. We summarized performance using hit rate@$K$ (the fraction of cases with at least one trial judged Recommended in the top $K$), recommended precision@$K$ (the fraction of top-$K$ recommendations judged Recommended), and eligible precision@$K$ (the fraction of top-$K$ recommendations judged Recommended or Eligible but Not Recommended). We designated hit rate@K the primary metric because it best reflects the intended workflow, in which a short ranked list should contain at least one clinician-recommended option.

Across these four cohorts, clinicians reviewed 440 patient-trial pairs from 188 de-identified cases spanning short call intake notes, brief advocacy summaries, concise tumor-board referrals, and longer consultation notes. Hit rate was already 0.88 at $K=1$, indicating that the top-ranked recommendation was clinician-recommended in 88\% of cases. Hit rate reached 0.91 at $K=3$ and remained at 0.91 through $K=10$, indicating that the evaluated top-10 list contained at least one clinician-recommended trial in 91\% of cases. Eligible precision remained high across cutoffs, decreasing only slightly from 0.89 at $K=1$ to 0.87 at $K=10$. Recommended precision, which required clinicians to judge a trial appropriate to pursue rather than merely eligible, was 0.88 at $K=1$ and 0.83 at $K=10$ (Fig.~\ref{fig:retrospective}b).

For the National Cancer Institute (NCI) cohort, comprising de-identified oncology referral notes from the Center for Immuno-Oncology (CIO) of the Center for Cancer Research (CCR), the predefined search space comprised a focused institutional portfolio of nine oncology trials, enabling exhaustive review of all nine trials for each of the 100 cases and yielding 900 patient-trial pairs. For the agreement analysis, the original clinician labels were mapped to the three TrialGPT 2.0 output categories as described in Methods. Each pair received three human annotations: one from an independent full-cohort review and two from a counterbalanced experiment in which two clinicians each reviewed all pairs, with TrialGPT 2.0 assistance assigned to opposite 50-case halves of the cohort. The resulting clinician consensus labels served as the reference standard.

Against these reference labels, TrialGPT 2.0 alone reached 89.0\% exact agreement, whereas clinician review with and without TrialGPT 2.0 reached 94.0\% and 94.2\%, respectively. TrialGPT 2.0 correctly identified 103 of 118 Highly Recommended pairs and 671 of 723 Low Fit pairs, corresponding to class-specific recalls of 87.3\% and 92.8\%, respectively. Agreement was lower for Possible Match pairs (27 of 59), consistent with the greater ambiguity of intermediate cases, for which incomplete information can affect adjudication. Although clinician review with and without assistance achieved similar overall agreement, the error profiles differed. Severe over-recommendation, defined as labeling a consensus-reference Low Fit pair as Highly Recommended, occurred in 1 pair with TrialGPT 2.0-assisted review, compared with 14 without assistance and 8 for TrialGPT 2.0 alone. Across both directions, assisted review produced 5 severe reversals between Highly Recommended and Low Fit, compared with 16 without assistance and 11 for TrialGPT 2.0 alone. This pattern suggests that TrialGPT 2.0 assistance may reduce clinically concerning reversals while preserving high agreement with consensus labels (Fig.~\ref{fig:retrospective}c).

Finally, for the NCI/CCR/CIO cohort, we assessed clinician screening time in the same counterbalanced experiment, defined as clinician-recorded trial-level decision time summed across the nine candidate trials (excluding patient-context review and automated TrialGPT 2.0 PDF report generation). TrialGPT 2.0 assistance reduced mean screening time per patient from 129.8 to 58.4 seconds, a 55.0\% reduction, with consistent effects in both groups (100.9 to 41.2 seconds; 158.8 to 75.5 seconds). Mean screening time did not differ significantly between the two clinicians (88.2 versus 100.0 seconds per patient), indicating that the reduction was not explained by a clinician-level speed difference. Overall, these results indicate that TrialGPT 2.0 can identify clinician-recommended options across heterogeneous clinical notes, improve screening efficiency, and maintain high agreement with expert consensus labels (Fig.~\ref{fig:retrospective}d).

\subsection*{TrialGPT 2.0 substantially expands clinician-selected trial recommendations in prospective precision oncology review}

\begin{figure}[p]
  \centering
  \includegraphics[width=\textwidth]{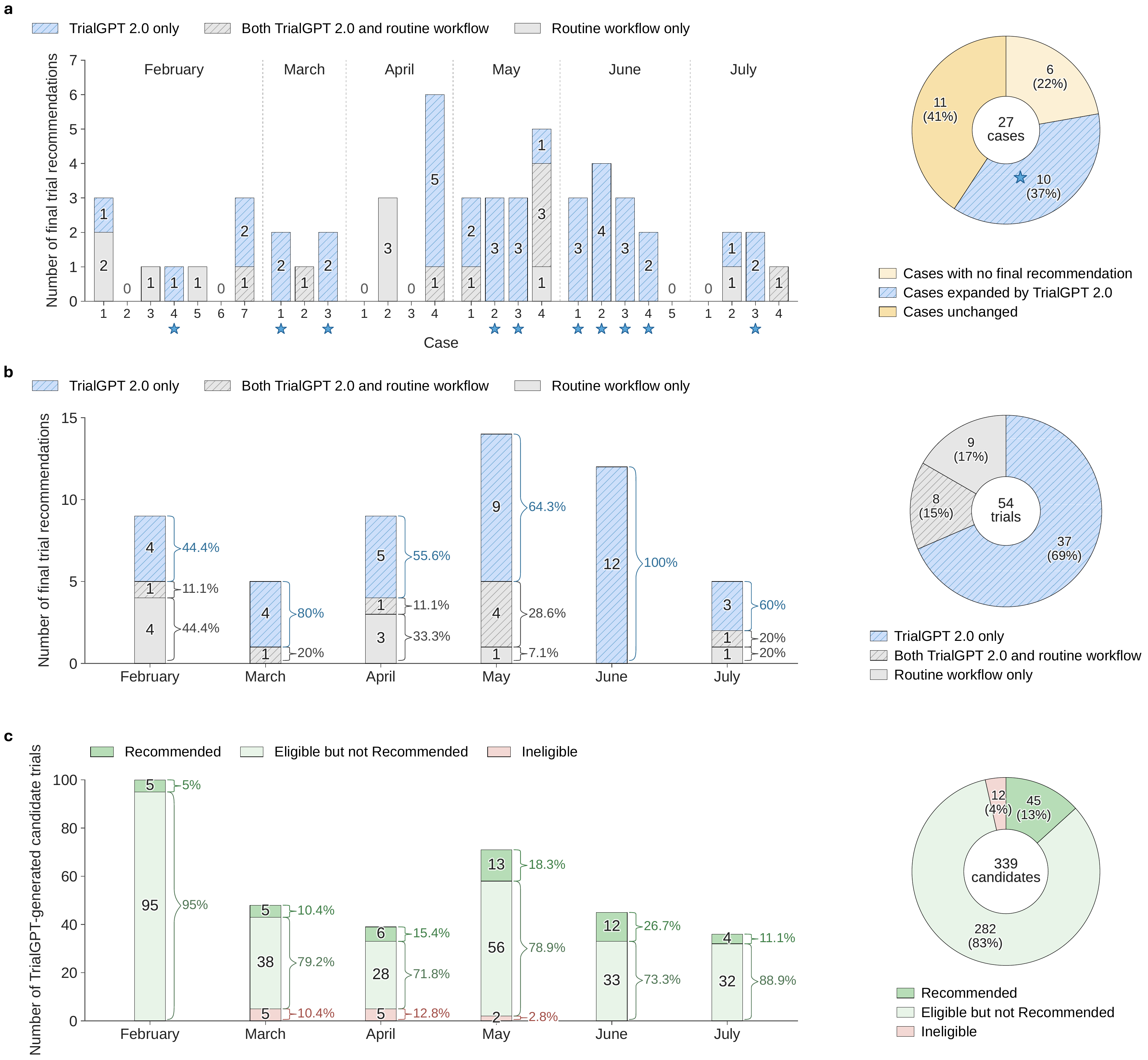}
  \caption{\textbf{Prospective evaluation of TrialGPT 2.0 in Precision Oncology Tumor Board (POTB) trial review.}
    \textbf{a,} Case-level source attribution of final clinician-selected trial recommendations across 27 POTB cases reviewed from February through July 2026. Bars show the number of final recommendations identified by TrialGPT 2.0 only, by both TrialGPT 2.0 and the routine workflow, or by the routine workflow only; cases without a final recommendation are shown as zero. The donut chart summarizes TrialGPT 2.0's case-level contribution: cases in which TrialGPT 2.0 expanded trial opportunities by contributing at least one final recommendation where the routine workflow identified none; cases in which opportunities were unchanged because the routine workflow had already identified at least one final recommendation; and cases with no final recommendation from either source. Stars mark the cases expanded by TrialGPT 2.0.
    \textbf{b,} Monthly and overall source attribution of the 54 final clinician-selected trial recommendations. Counts and percentages indicate recommendations identified by TrialGPT 2.0 only, by both TrialGPT 2.0 and the routine workflow, or by the routine workflow only.
    \textbf{c,} Monthly and overall clinician adjudication of the 339 candidate trials generated by TrialGPT 2.0. Candidate trials were labeled as Recommended, Eligible but Not Recommended or Ineligible. Recommended indicates selection for the final POTB recommendation list; Eligible but Not Recommended indicates a trial considered eligible but not prioritized for final recommendation; and Ineligible indicates incompatibility with the patient context or trial criteria.
  }
  \label{fig:prospective}
\end{figure}

We prospectively evaluated TrialGPT 2.0 in the UIC POTB, a routine multidisciplinary precision-oncology workflow that reviews clinical context, molecular findings, prior therapies, and the clinical question to identify relevant trial options. Cases entered through routine POTB referral. Trial identification proceeded in parallel: clinical reviewers identified candidate trials through the routine workflow, while TrialGPT 2.0 generated recommendations from the same case context. The two lists were reviewed together, and clinicians selected the final POTB recommendation list.
The primary endpoint was TrialGPT 2.0's contribution to the final clinician-selected recommendations, assessed at the case level (whether TrialGPT 2.0 contributed at least one final clinician-selected recommendation in cases where the routine workflow found none) and at the recommendation level (the source of each final recommendation) (Fig.~\ref{fig:prospective}a,b). The secondary endpoint characterized all TrialGPT 2.0-generated candidate trials using clinician labels (Fig.~\ref{fig:prospective}c).

At the case level, the analysis included 27 POTB cases reviewed from February through July 2026 (Fig.~\ref{fig:prospective}a). TrialGPT 2.0 identified final recommendations in 10 cases for which the routine workflow had found none, increasing the number of cases with at least one final recommendation from 11 to 21, a 90.9\% increase. Considering all 27 reviewed cases, TrialGPT 2.0 expanded final trial options in 10 cases (37.0\%), left opportunities unchanged in 11 (40.7\%) where the routine workflow had already identified at least one recommendation, and neither source contributed a final recommendation in 6 cases (22.2\%).

At the recommendation level, clinicians selected 54 final trial recommendations, of which TrialGPT 2.0 identified 45 (83.3\%): 37 (68.5\%) by TrialGPT 2.0 alone and 8 (14.8\%) by both TrialGPT 2.0 and the routine workflow (Fig.~\ref{fig:prospective}b). The routine workflow identified 17 recommendations in total (the 8 shared with TrialGPT 2.0 and 9 identified by the routine workflow alone). The 37 recommendations uniquely contributed by TrialGPT 2.0 therefore more than doubled the final recommendation set, expanding it from 17 to 54.

For the secondary endpoint, we analyzed clinician labels for the 339 candidate trials TrialGPT 2.0 generated across the 27 cases (Fig.~\ref{fig:prospective}c). UIC clinicians labeled 45 (13.3\%) as Recommended, 282 (83.2\%) as Eligible but Not Recommended, and 12 (3.5\%) as Ineligible. The predominance of Eligible but Not Recommended labels directly illustrates why eligibility alone is insufficient in precision oncology: most candidate trials were not clearly incompatible, yet only a small subset was prioritized for the final recommendation list. Among high-scoring candidates labeled Eligible but Not Recommended, clinicians deprioritized technically eligible trials when poor performance status or comorbidities made intensive regimens less appropriate, when standard-of-care or more specific targeted options were preferred, when biomarker evidence was limited, or when the POTB question was diagnostic rather than therapeutic. These patterns indicate that clinical recommendation in precision oncology requires expert integration of treatment intent, expected benefit, molecular relevance, evidence strength, feasibility, and patient-specific context beyond protocol eligibility alone.

As an exploratory failure-mode analysis, we reviewed the three cases in which all final recommendations came from the routine workflow alone. In each, TrialGPT 2.0 applied protocol eligibility criteria literally, whereas clinicians interpreted the same information using broader oncologic context. In one case, TrialGPT 2.0 excluded a patient from a first-line trial because the record showed prior chemotherapy; clinicians recognized that this therapy had been given when the patient's lung lesions were initially misclassified as metastases from another primary, leaving the lung tumor itself effectively untreated. In another, TrialGPT 2.0 rejected a strong biomarker-matched trial because the protocol required progression on a specific prior drug with no intervening treatment, whereas the patient had received an interim therapy. In the third, TrialGPT 2.0 down-ranked biologically relevant trials because poor performance status and severe comorbidities, including advanced dementia, raised eligibility concerns. In each case, clinicians drew on specialist judgment about tumor biology, attribution of prior treatment, and trial feasibility to retain these trials as reasonable options for discussion.

\subsection*{NIH-TrialBench supports reproducible benchmarking for AI-assisted clinical trial matching}

\begin{figure}[p]
  \centering
  \includegraphics[width=\textwidth]{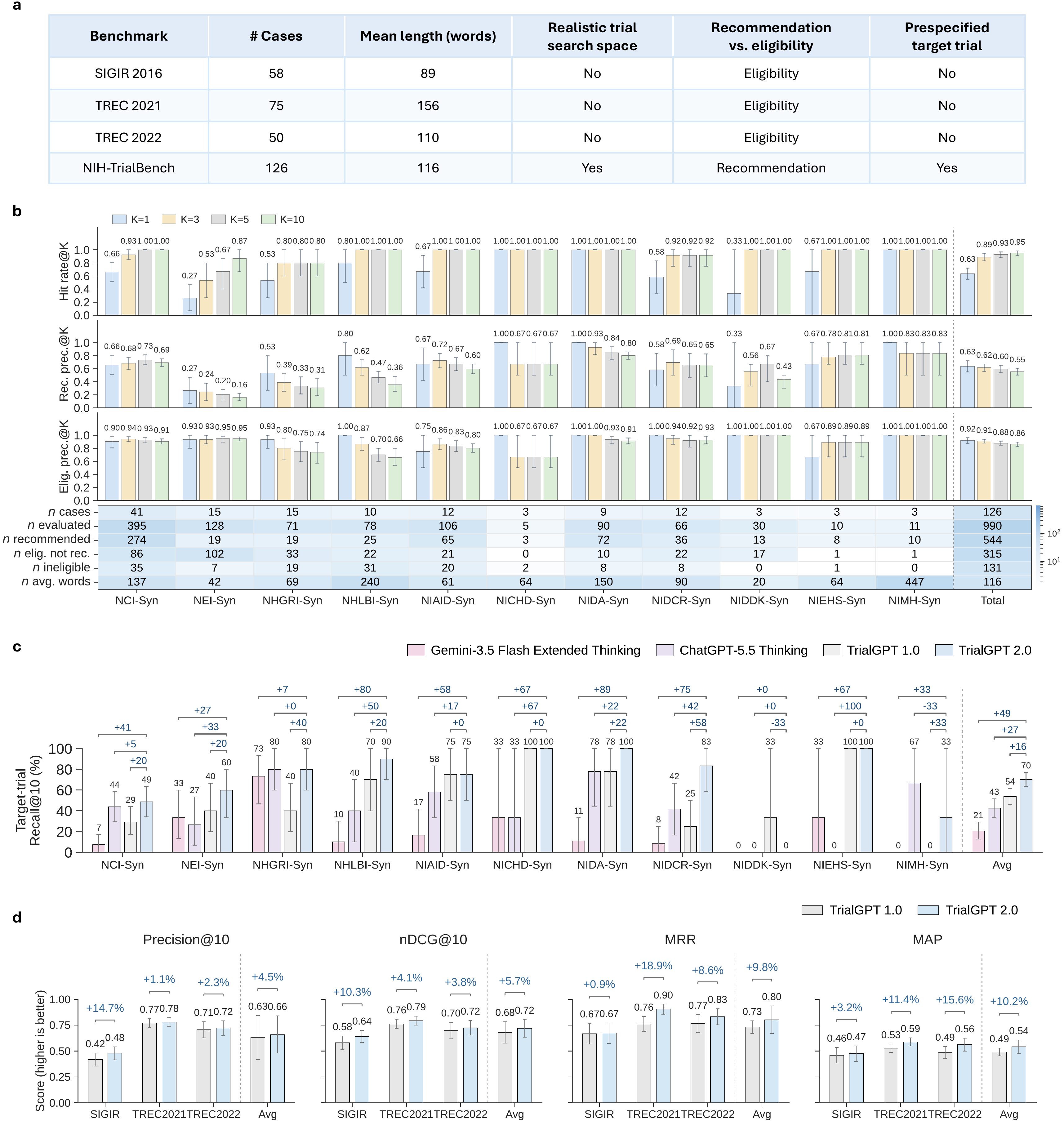}
  \caption{\textbf{TrialGPT 2.0 performance in reproducible benchmarking settings.}
    \textbf{a,} Comparison of NIH-TrialBench with established public benchmarks across characteristics.
    \textbf{b,} NIH-TrialBench top-ranked review performance. Bars show hit rate@$K$, recommended precision@$K$, and eligible precision@$K$ for $K = 1, 3, 5$ and $10$, with error bars showing 95\% confidence intervals of the mean computed by percentile bootstrap (10{,}000 resamples). The heatmap summarizes the number of vignettes, reviewed patient-trial pairs, clinician labels and average vignette length for each contributing NIH Institute or Center and overall. 
    \textbf{c,} Target-trial Recall@10 on NIH-TrialBench, comparing TrialGPT 2.0 with TrialGPT 1.0 and off-the-shelf general-purpose models, shown per contributing NIH Institute or Center and as the mean across institutes. Error bars show 95\% confidence intervals computed by percentile bootstrap (10{,}000 resamples), and annotations indicate the percentage-point difference between TrialGPT 2.0 and each comparator model.
    \textbf{d,} Public benchmark comparison. TrialGPT 2.0 is compared with TrialGPT 1.0 using Precision@10, nDCG@10, MRR and MAP. Bars show mean performance, error bars show approximate 95\% confidence intervals (mean $\pm$ 1.96 $\times$ standard error across patient queries) and percentage annotations indicate relative change for TrialGPT 2.0 compared with TrialGPT 1.0.
  }
  \label{fig:synthetic}
\end{figure}

To complement the retrospective and prospective evaluations, we assessed TrialGPT 2.0 in two reproducible benchmarking settings: NIH-TrialBench and established public benchmarks. Unlike existing eligibility-oriented public benchmarks, NIH-TrialBench evaluates clinician-adjudicated labels against a realistic, workflow-defined trial search space and includes a prespecified target trial for each vignette, enabling evaluation of clinical recommendation quality and target-trial recovery (Fig.~\ref{fig:synthetic}a). 

On NIH-TrialBench, using the same top-ranked review framework as the retrospective cohorts (Fig.~\ref{fig:retrospective}b), clinicians reviewed up to ten top-ranked trials categorized by TrialGPT 2.0 as Highly Recommended for each vignette and labeled each patient-trial pair as Recommended, Eligible but Not Recommended, or Ineligible. Across 126 vignettes and 990 reviewed patient-trial pairs, eligible precision was 0.92 at $K = 1$ and 0.86 at $K = 10$, and hit rate increased from 0.63 at $K = 1$ to 0.95 at $K = 10$, indicating that the top-10 recommendation list included at least one clinician-recommended trial for most vignettes (Fig.~\ref{fig:synthetic}b).

Because each vignette was constructed around a known intended target trial, NIH-TrialBench also enabled direct assessment of target-trial recovery, measured as Recall@10 (the proportion of vignettes for which the known target trial appeared among the ten highest-ranked trials). TrialGPT 2.0 achieved a macro-averaged target-trial Recall@10 of 70\% across the 11 contributing NIH Institutes and Centers, compared with 54\% for the original TrialGPT framework~\cite{jin2024matching} (hereafter TrialGPT 1.0) and, among off-the-shelf general-purpose models, 43\% for ChatGPT-5.5 Thinking~\cite{openai2026gpt55} and 21\% for Gemini-3.5 Flash Extended Thinking~\cite{google2026gemini35}, indicating stronger recovery of intended matches by the trial-specific recommendation system (Fig.~\ref{fig:synthetic}c). The recommendation category assigned to each target trial is summarized in Supplementary Fig.~\ref{fig:supplementary-synthetic-vignettes}.

Finally, on the public benchmarks (SIGIR and the TREC 2021 and TREC 2022 Clinical Trials tracks~\cite{koopman2016test,soboroff2021overview,roberts2022overview}), which provide standardized relevance judgments for trial-ranking evaluation, TrialGPT 2.0 improved ranking performance over TrialGPT 1.0 across all four metrics (Fig.~\ref{fig:synthetic}d). Averaged across the three benchmarks, TrialGPT 2.0 improved Precision@10 by 4.5\%, normalized discounted cumulative gain at 10 (nDCG@10) by 5.7\%, mean reciprocal rank (MRR) by 9.8\% and mean average precision (MAP) by 10.2\%. In addition, it reduced inference burden, increasing processing speed by a factor of 3.20 and reducing input and output token counts per patient-trial pair by 58\% and 73\%, respectively (\hyperref[suppnote:public-benchmark]{Supplementary Note 4} and Supplementary Fig.~\ref{fig:supplementary-backbone}b). Together, these results complement the real-world evaluations by demonstrating improved recommendation-oriented performance, ranking quality, and computational efficiency under reproducible benchmark conditions.

\subsection*{Real-world deployment, runtime and usability of TrialGPT 2.0}

Finally, we deployed TrialGPT 2.0 as a web-based interface for trial recommendation review across participating oncology and referral workflows. The interface accepts a patient clinical summary and uses the corresponding predefined local trial corpus and matching policy to generate ranked recommendations, displayed as trial cards grouped into Highly Recommended, Possible Match or Low Fit. Each trial card presents trial metadata and a structured rationale, including eligibility-supporting evidence, missing information and ineligibility concerns. Clinical reviewers and recruitment specialists can inspect recommendations by category, apply post-ranking filters and export matched results for review (Supplementary Fig.~\ref{fig:interface}).

In runtime testing across 25 cases sampled from the five participating settings, TrialGPT 2.0 generated recommendations in 21.7 seconds on average, including retrieval when applicable, backend assessment, output parsing, result writing and score-sorted PDF generation. Runtime varied across settings, reflecting differences in case length, trial-corpus size and number of matched trials, but remained compatible with rapid generation of candidate trial lists for clinical review. In practical deployment, the interface supports recommendation generation from a nationwide corpus of more than 25{,}000 active clinical trials with sites in the United States. Additional runtime details are provided in \hyperref[suppnote:interface-time]{Supplementary Note 5} and Supplementary Fig.~\ref{fig:supplementary-time-cost}.

Clinical reviewer acceptance was assessed using an 8-item questionnaire covering usability, workflow fit, perceived efficiency, clarity, decision confidence, usefulness, content safety, and overall satisfaction on a 5-point Likert scale. Overall, responses were favorable, with 85 of 104 item-level ratings scored as 4 or 5. Content safety received the highest mean rating of 4.92, with all responses favorable. Overall satisfaction and usability each received a mean score of 4.54, followed by clarity and workflow fit, with mean scores of 4.38 and 4.31, respectively. Ratings were more variable for perceived efficiency and decision confidence, with mean scores of 3.85 and 3.92, respectively, suggesting that perceived benefit may depend on workflow context, reviewer role, and continued user calibration. These findings support the operational feasibility of TrialGPT 2.0 across heterogeneous oncology and referral workflows, while highlighting the need for continued workflow integration and reviewer training during deployment (\hyperref[suppnote:reviewer-acceptance]{Supplementary Note 6}, Supplementary Table~\ref{tab:clinical_reviewer_acceptance} and Supplementary Fig.~\ref{fig:supplementary-reviewer-acceptance-responses}).

\section*{Discussion}

In this multicenter study, we present TrialGPT 2.0, a real-world system designed to support responsible AI-assisted patient-trial recommendation, and evaluate it through retrospective review, prospective precision-oncology tumor-board assessment, reproducible benchmarking, and deployment. Across retrospective cohorts, TrialGPT 2.0 identified clinician-recommended trials within short ranked lists and reduced clinician screening time while preserving high agreement with consensus reference labels. 
In the prospective UIC POTB workflow, it expanded trial opportunities by identifying final recommendations for cases whose routine review yielded no final recommendation.
On NIH-TrialBench and public benchmarks, it improved target-trial recovery and ranking performance relative to TrialGPT 1.0.
TrialGPT 2.0 has been deployed through a secure web-based interface for trial review in real-world clinical workflows. The web-based interface generated recommendations within a short turnaround and received generally favorable ratings from clinical reviewers, supporting the operational feasibility of AI-assisted trial matching in routine trial-review workflows.

These findings address two gaps that have limited the clinical value of prior AI-assisted trial matching. First, as in much of medical AI, prior trial-matching methods have been evaluated largely using curated benchmarks that measure technical performance rather than real-world clinical contribution~\cite{showus2026,kann2021artificial}. By instead assessing TrialGPT 2.0 on real clinical data through clinician adjudication and prospective review in an active precision-oncology workflow, we show that AI-identified trials can contribute to real clinical decisions. To make such evaluation reproducible under the privacy constraints on real clinical notes, we developed and released NIH-TrialBench, which pairs clinician-authored synthetic vignettes built around known target trials with clinician-adjudicated labels and a realistic, workflow-defined trial search space.

Second, identifying eligible trials is not sufficient for clinical recommendation. Across the evaluations, reviewers explicitly separated trials that were Recommended from those that were Eligible but Not Recommended, reflecting differences in clinical appropriateness and priority that eligibility assessment alone does not capture. 
Accordingly, TrialGPT 2.0 produces graded recommendation categories and a fit score rather than a flat set of eligible trials, jointly assessing formal eligibility, patient-trial fit, and workflow-specific clinical priorities. Its fit score responds to clinically important mismatches in diagnosis or histology, disease status, biomarker or pathway match, line of therapy, prior treatment, and cohort or arm applicability, and is higher for clinician-recommended trials (Supplementary Fig.~\ref{fig:validity}). Importantly, clinician adjudication showed that technically eligible trials may still be deprioritized because of treatment intent, limited expected benefit, preference for standard-of-care or more specific targeted options, treatment-sequence conflicts, feasibility, or patient-specific clinical context. Patient-trial matching is therefore better framed and evaluated as a recommendation task that extends beyond eligibility classification.

Beyond these clinical contributions, TrialGPT 2.0's design, evaluation, and governance address all seven ethical elements identified for AI-enabled clinical trial recruitment~\cite{rentzepis2026use}. By embedding these ethical considerations into the system architecture, evaluation framework, and deployment model, TrialGPT 2.0 combines clinical utility with responsible AI safeguards.

This study has several limitations. 
First, we evaluated early-stage trial-matching outcomes, including recommendation retrieval, clinician selection, agreement and screening time, but did not assess downstream recruitment outcomes such as patient contact, referral, formal eligibility screening, consent, enrollment, accrual or patient outcomes. These downstream processes require longer follow-up and are influenced by multiple factors beyond trial matching, including patient preferences, changes in clinical status, trial availability, logistical barriers and site-level recruitment procedures. Assessing the effect of TrialGPT 2.0 on actual enrollment and patient outcomes was therefore beyond the scope of this study. Nonetheless, our findings complement and substantially extend the limited existing real-world evidence, which has largely been derived from narrowly scoped evaluations of discrete trial-matching tasks and constrained clinical settings~\cite{unlu2025manual,gueguen2025prospective,syed2026translating,parikh2026human}, and provide important evidence that AI-enabled trial matching can meaningfully contribute to improved patient recruitment and access to clinical trials.
In addition, the prospective evaluation was conducted in a single POTB workflow over a limited period with a modest number of cases; larger prospective evaluations across institutions, disease programs and recruitment workflows are needed to assess generalizability and downstream impact~\cite{liu2020reporting,rivera2020guidelines}, ideally measuring trials discussed with patients, screening referrals, enrollment outcomes and accrual efficiency~\cite{showus2026}. Encouragingly, the original TrialGPT framework~\cite{jin2024matching} has already been independently deployed for real-world eligibility screening at an external institution~\cite{syed2026translating}, supporting the feasibility of adapting the approach outside its original development setting.
Second, the evaluated trial corpora and clinical workflows were predominantly U.S.-based. International deployment would require integrating additional regional and international trial data sources, such as the EU Clinical Trials Information System~\cite{ema_ctis}, the WHO International Clinical Trials Registry Platform~\cite{who_ictrp} and the Japan Registry of Clinical Trials~\cite{jrct}. Because TrialGPT 2.0 is designed to be transferable across standardized candidate trial corpora, incorporating such sources is feasible and would broaden the system's applicability beyond the settings evaluated here.
Third, the main evaluation used GPT-4.1 as the backbone model, which was available when system development and evaluation began but may be deprecated over time. Since TrialGPT 2.0 can be interpreted as a versioned AI-assisted system rather than a single static model, the backbone can be replaced with newer or alternative models. Supplementary analyses show that it transfers to GPT-5.4 and Gemini-3.1-pro-preview with a similar overall pattern of performance (Supplementary Fig.~\ref{fig:supplementary-backbone}).
Fourth, error analysis revealed failure modes in both directions. TrialGPT 2.0 occasionally applied eligibility criteria too literally, missing trials that clinicians retained as reasonable options by drawing on broader oncologic context, such as the clinical interpretation of prior treatment or trial feasibility (Fig.~\ref{fig:prospective}). Conversely, among trials that TrialGPT 2.0 rated Highly Recommended but clinicians did not recommend, discordances reflected condition, biomarker or protocol mismatches, as well as clinical-priority concerns such as limited expected benefit or treatment-sequence conflict (Supplementary Fig.~\ref{fig:supplementary-errors}). Reducing these residual errors, particularly in complex multi-arm trials and cases with incomplete or clinically ambiguous information, is an important direction for future work.

Our findings show that AI-assisted matching can identify additional trial options that clinicians select for final recommendation. Realizing its full value will require integration into rigorous clinical workflows, ongoing ethical and performance oversight, and large-scale, outcome-based evaluation. With continued monitoring, local revalidation, and studies that measure patient and access outcomes, AI-assisted matching could progress from suggesting candidate trials to producing measurable improvements in trial access, helping more patients reach appropriate trials earlier and more efficiently.

\newpage
\section*{Methods}

\subsection*{TrialGPT 2.0 system design}

TrialGPT 2.0 takes three inputs for each matching task: patient context, a predefined local trial corpus and a local matching policy. Patient context can be provided as raw clinical records or as a clinician-prepared narrative summary. The local trial corpus comprises the predefined set of candidate trials to be considered for matching in a given recruitment setting, including trial metadata, eligibility criteria and cohort or arm information. The local matching policy specifies clinical preferences that guide matching and ranking, including clinical relevance weighting, disease- or biomarker-specific priorities and therapeutic trial preference. These preferences guide prioritization but do not override explicit eligibility criteria.

TrialGPT 2.0 uses a pluggable retrieval component to support real-time matching across trial corpora of different sizes. For large trial pools, detailed backend assessment of every trial can be computationally inefficient and unsuitable for real-time review. In these settings, the default retriever follows the hybrid-fusion retrieval strategy of the original TrialGPT~\cite{jin2024matching} framework. Briefly, the retriever generates patient-derived search queries from the patient context, applies both lexical and semantic retrieval to the local trial corpus, and merges query-level results using reciprocal rank fusion to produce a candidate shortlist. In our implementation, lexical retrieval is based on BM25~\cite{robertson2009probabilistic} and semantic retrieval is based on MedCPT~\cite{jin2023medcpt}, consistent with the original TrialGPT retrieval design. For smaller trial pools, defined in this study as fewer than 1{,}500 candidate trials, TrialGPT 2.0 passes the full predefined trial corpus directly to the backend. The retrieval component determines which trials proceed to detailed assessment, whereas the backend assigns the final fit score, confidence estimate, recommendation category and rank.

The TrialGPT 2.0 backend generates a single trial-level assessment for each candidate trial. When a trial includes multiple cohorts or arms, the backend evaluates whether the patient fits any clinically meaningful option and produces one overall recommendation for the trial. The assessment jointly considers formal eligibility, clinical relevance and uncertainty in the available patient information. Clear violations of required eligibility criteria lower the fit score and appear as ineligible reasons, whereas missing information that could typically be obtained during screening lowers confidence rather than automatically indicating ineligibility. When line-of-therapy information is available, the backend uses it to determine whether the patient's prior treatment history matches the treatment setting required by the trial or relevant cohort.

For each candidate trial, TrialGPT 2.0 returns a strict structured output containing eligible reasons, missing information, ineligible reasons, rationale, fit score and confidence estimate. Eligible reasons summarize criteria that the patient appears to meet; ineligible reasons summarize explicit mismatches or exclusion concerns; missing information captures key unknowns that affect assessment; and the rationale provides an overall explanation of the recommendation. The fit score is an integer from 0 to 100 and represents the overall strength of patient-trial alignment. Recommendation categories are assigned from the fit score: scores greater than 90 are classified as Highly Recommended, scores from 80 to 90 as Possible Match and scores below 80 as Low Fit. The confidence estimate ranges from 0 to 1 and reflects the reliability of the assessment, accounting for missing or uncertain information. The final rank orders trials primarily by fit score; when trials have the same fit score, confidence serves as a tie-breaking criterion, with higher-confidence assessments ranked above lower-confidence assessments.

Unless otherwise specified, all main analyses used GPT-4.1~\cite{openai2025gpt41}, accessed through the Azure OpenAI Service, as the backbone model. Model decoding settings, code version, prompt and backbone model were fixed across evaluations. Analyses with alternative backbone models (GPT-5.4~\cite{openai2026gpt54} and Gemini-3.1-pro-preview~\cite{googleai2026gemini31propreview}) are reported in Supplementary Fig.~\ref{fig:supplementary-backbone}.

Matching policies were developed with clinician input to reflect the review priorities and available patient information in each recruitment or evaluation setting. Consequently, the trial-fit factors represented in the policies, and the depth with which they were assessed, varied across workflows (\hyperref[suppnote:factor]{Supplementary Note 1} and Supplementary Table~\ref{tab:factor_coverage}). For the NIH-TrialBench evaluation, TrialGPT 2.0 uses a general matching policy that is fixed before evaluation and is not refined using NIH-TrialBench vignettes or annotations. For retrospective cohorts, local matching policies are refined using clinician feedback from cases other than the case being evaluated. For each retrospective evaluation case, the policy is fixed before inference, and any clinician feedback or error analysis from that same case is not used to revise the policy that generated its reported result.

Because LLM backbones, trial registries and local trial portfolios can change over time, we evaluated TrialGPT 2.0 as a versioned AI-assisted system. For NIH-TrialBench, retrospective clinical-note cohorts, public benchmark analyses and runtime experiments, the code version, prompt version, backbone model, decoding settings, retrieval configuration, trial-corpus or benchmark candidate set and matching policy were fixed before inference. For the prospective POTB evaluation, the system used the current weekly refreshed UIC trial-corpus snapshot available at the time of each case review, while the code version, prompt version, backbone model, decoding settings, retrieval configuration and matching policy were recorded. Reported results therefore apply to the TrialGPT 2.0 configurations and trial-corpus snapshots used for each evaluation. Substantive changes to the backbone model, prompts, retrieval module, trial corpus or matching policy should be treated as a new system version and undergo local revalidation before clinical use.

\subsection*{Study cohorts}

The study included three cohort types: de-identified retrospective clinical notes, prospective Precision Oncology Tumor Board (POTB) cases and clinician-authored synthetic vignettes. These cohort types supported complementary evaluation of TrialGPT 2.0 across real clinical documentation, an active clinical trial-review workflow and reproducible benchmark conditions.

The retrospective evaluation used 288 de-identified cases from five recruitment pathways: patient call intake notes from the NIH Office of Patient Recruitment (OPR), oncology patient advocacy intake notes from the Cholangiocarcinoma Foundation (CCF), tumor-board referrals from the University of Illinois Cancer Center (UIC), oncology consultation notes from the University of Pittsburgh Medical Center (UPMC) and oncology referral notes from the National Cancer Institute (NCI). For each retrospective cohort, the contributing site or program defined a candidate trial corpus before TrialGPT 2.0 inference. These corpora reflected the trials considered relevant to that recruitment setting and were fixed during retrospective inference and clinician review. Trial search-space size varied across cohorts, from 9 to 1{,}871 candidate trials.

The prospective evaluation was conducted in the UIC Precision Oncology Tumor Board (POTB) workflow. Prospective cases entered the workflow through routine tumor-board referral and were reviewed using de-identified case context, including clinical history, molecular findings, prior therapies, and the clinical question. Unlike the fixed retrospective trial-corpus snapshots, the prospective workflow used the current UIC trial-corpus snapshot available at the time of each case review, reflecting weekly updates to trial availability and recruitment status.

To support reproducible evaluation, we created NIH-TrialBench, a clinician-authored synthetic benchmark based on actual NIH trials. NIH-TrialBench was developed from target-trial-informed vignettes contributed by 24 investigators, with target trials spanning 11 NIH Institutes and Centers. The main NIH-TrialBench evaluation set included 126 synthetic vignettes and a predefined corpus of 1{,}373 candidate trials from the NIH Clinical Center's \textit{Search the Studies} registry~\cite{nih_search_studies}. Each vignette corresponded to one target trial, defined as the intended match for the synthetic patient. Because the vignettes are synthetic, NIH-TrialBench supports reproducible benchmarking without exposing private patient information.

Cohort-specific review strategies are summarized in Fig.~\ref{fig:cohorts}a. Top-ranked review was used for NIH-TrialBench and the OPR, CCF, UIC and UPMC retrospective cohorts. Exhaustive review was used for the NCI retrospective cohort because each case had only nine predefined candidate trials, making complete review feasible. Prospective review was used in the UIC POTB workflow, where TrialGPT 2.0-generated candidate trials were reviewed during ongoing trial recommendation. 

We manually reviewed clinician annotations for quality control before analysis, including checks for completeness and consistency between annotation comments and assigned labels. Annotation issues identified during quality control were resolved before metric calculation.

We characterized disease-area coverage using first-level disease categories from Medical Subject Headings (MeSH)~\cite{lowe1994understanding}. For each retrospective clinical note, prospective POTB case and NIH-TrialBench vignette, GPT-5.4~\cite{openai2026gpt54}, accessed through the Azure OpenAI Service, mapped the patient context to up to three predefined MeSH disease categories, ordered by relevance. Model outputs were restricted to the predefined category list and manually reviewed for quality control. We summarized disease-area coverage as the number of cases in each cohort assigned to each MeSH disease category (Fig.~\ref{fig:cohorts}b). Age group and sex were obtained from case metadata or vignette text when available; sex was recorded as unknown when not specified (Fig.~\ref{fig:cohorts}c).

\subsection*{Retrospective evaluation}

Retrospective clinical-note evaluation assessed whether TrialGPT 2.0 recommendations aligned with clinician judgment and whether TrialGPT 2.0 assistance reduced clinician screening time. The retrospective analysis included two review strategies. Top-ranked review was applied to the OPR, CCF, UIC, and UPMC cohorts and evaluated whether highly ranked TrialGPT 2.0 recommendations were judged eligible or recommended by clinicians. Exhaustive review was applied to the NCI cohort, where each case had a predefined search space of nine candidate trials, enabling complete-pool agreement analysis and a counterbalanced clinician-assistance experiment.

For top-ranked review, TrialGPT 2.0 ranked the predefined candidate trial corpus for each case. Clinicians reviewed up to ten highest-ranked trials categorized by TrialGPT 2.0 as Highly Recommended and assigned each reviewed patient-trial pair one of three labels: Recommended, Eligible but Not Recommended, or Ineligible. Recommended indicates that the patient appears eligible and that the trial is clinically advisable for consideration. Eligible but Not Recommended indicates that the patient appears technically eligible but that the trial is not clinically advisable or not a priority in the current clinical context. Ineligible indicates that the patient does not meet trial eligibility criteria or has a clear incompatibility with the trial. We evaluated top-ranked review performance at \(K \in \{1,3,5,10\}\). For these metrics, 95\% confidence intervals were computed by percentile bootstrap over cases with 10{,}000 resamples (Fig.~\ref{fig:retrospective}b).

Let \(q\) denote a case and let \(\mathcal{Q}_{\mathrm{top}}\) denote the set of retrospective cases in the top-ranked review analysis. Let \(S_q(K)\) denote the first \(\min(K,n_q)\) annotated trials for case \(q\), where \(n_q\) is the number of annotated Highly Recommended trials for that case. Let \(y_{q,t}\) denote the clinician label for trial \(t\) in case \(q\), with
\begin{equation}
y_{q,t} \in \{\mathrm{Recommended},\ \mathrm{Eligible\ but\ Not\ Recommended},\ \mathrm{Ineligible}\}.
\end{equation}
We define the set of clinically eligible labels as
\begin{equation}
\mathcal{E}=\{\mathrm{Recommended},\ \mathrm{Eligible\ but\ Not\ Recommended}\}.
\end{equation}

Hit rate@$K$ measures case-level success in identifying at least one clinician-recommended trial within the evaluated top-\(K\) list:
\begin{equation}
\mathrm{Hit\ rate@}K
=
\frac{1}{|\mathcal{Q}_{\mathrm{top}}|}
\sum_{q \in \mathcal{Q}_{\mathrm{top}}}
\mathbb{I}
\left(
\exists\, t \in S_q(K): y_{q,t}=\mathrm{Recommended}
\right).
\end{equation}
Because clinician review was restricted to model-selected Highly Recommended trials, hit rate@$K$ measures whether the evaluated top-$K$ list contained at least one clinician-recommended trial and should not be interpreted as recall over the full candidate trial corpus.

\begin{equation}
\mathrm{Recommended\ precision@}K
=
\frac{1}{|\mathcal{Q}_K|}
\sum_{q \in \mathcal{Q}_K}
\frac{1}{|S_q(K)|}
\sum_{t \in S_q(K)}
\mathbb{I}\left(y_{q,t}=\mathrm{Recommended}\right).
\end{equation}

Eligible precision@$K$ measures the fraction of evaluated top-\(K\) recommendations judged clinically eligible by clinicians:
\begin{equation}
\mathrm{Eligible\ precision@}K
=
\frac{1}{|\mathcal{Q}_K|}
\sum_{q \in \mathcal{Q}_K}
\frac{1}{|S_q(K)|}
\sum_{t \in S_q(K)}
\mathbb{I}\left(y_{q,t} \in \mathcal{E}\right),
\end{equation}
where \(\mathcal{Q}_K=\{q:|S_q(K)|>0\}\).

The NCI cohort represents a distinct evaluation setting because clinicians exhaustively reviewed every trial in a small, curated portfolio of nine potentially relevant oncology trials. Since the search space had already been narrowed to trials relevant to the workflow, the principal task was eligibility triage. \textit{Eligible} indicated that no current eligibility barrier had been identified and that the trial could proceed to further consideration or screening. \textit{Subject-to-change ineligible} indicated that eligibility could change with additional information or modification of a clinical factor, whereas \textit{Definitely ineligible} indicated a clear, non-resolvable incompatibility. For the three-class agreement analysis, we mapped \textit{Eligible} to TrialGPT 2.0’s Highly Recommended category, \textit{Subject-to-change ineligible} to Possible Match and \textit{Definitely ineligible} to Low Fit. In this focused setting, mapping \textit{Eligible} to Highly Recommended is not contradictory because Eligible functioned as the positive action category for proceeding to further consideration. This mapping was used only to align the original clinician labels with the system’s three output categories. We denote this label set as
\begin{equation}
\mathcal{C}=
\{\mathrm{Highly\ Recommended},\ \mathrm{Possible\ Match},\ \mathrm{Low\ Fit}\}.
\end{equation}

Each NCI patient-trial pair received three initial human annotations. 
Before the assistance experiment, an independent clinician blinded to TrialGPT 2.0 outputs annotated all 900 patient-trial pairs. The same pairs were then reviewed by two additional clinicians in a counterbalanced TrialGPT 2.0 assistance experiment. The 100 cases were divided into two groups of 50: one clinician reviewed the first group with TrialGPT 2.0 assistance and the second group without assistance, while the other clinician reviewed the groups under the opposite assistance assignments. Thus, each patient-trial pair received one independent blinded annotation, one annotation with TrialGPT 2.0 assistance, and one annotation without assistance. When all three initial annotations agreed, their shared label was used as the consensus reference label. When the annotations disagreed, a fourth clinician reviewed the discordant pair and made the final determination.

The labels resulting from this process were used as consensus reference labels for agreement analysis. Let \(r_{q,t}\) denote the consensus reference label for trial \(t\) in case \(q\). Let \(m_{q,t}\) denote the TrialGPT 2.0 recommendation category, \(h^{+}_{q,t}\) denote the clinician label assigned with TrialGPT 2.0 assistance, and \(h^{-}_{q,t}\) denote the clinician label assigned without TrialGPT 2.0 assistance, with
\begin{equation}
r_{q,t},m_{q,t},h^{+}_{q,t},h^{-}_{q,t} \in \mathcal{C}.
\end{equation}

We compared three assigned-label sources with the consensus reference labels: TrialGPT 2.0 alone, clinician review with TrialGPT 2.0 assistance, and clinician review without TrialGPT 2.0 assistance. For each source \(s\), let \(z^{(s)}_{q,t}\) denote its assigned label, where
\begin{equation}
z^{(s)}_{q,t} \in \{m_{q,t}, h^{+}_{q,t}, h^{-}_{q,t}\}.
\end{equation}
We computed a three-class confusion matrix for each source. Rows correspond to assigned labels and columns correspond to consensus reference labels:
\begin{equation}
C^{(s)}_{a,b}
=
\sum_{q \in \mathcal{Q}_{\mathrm{NCI}}}
\sum_{t \in \mathcal{T}_q}
\mathbb{I}\left(z^{(s)}_{q,t}=a,\ r_{q,t}=b\right),
\end{equation}
where \(a,b \in \mathcal{C}\), \(\mathcal{Q}_{\mathrm{NCI}}\) denotes the NCI retrospective cohort and \(\mathcal{T}_q\) denotes the nine candidate trials for case \(q\).

Exact agreement for source \(s\) is defined as the proportion of patient-trial pairs for which the assigned label matches the consensus reference label:
\begin{equation}
\mathrm{Agreement}^{(s)}
=
\frac{
\sum_{c \in \mathcal{C}} C^{(s)}_{c,c}
}{
\sum_{a \in \mathcal{C}}
\sum_{b \in \mathcal{C}}
C^{(s)}_{a,b}
}.
\end{equation}

Class-specific recall for reference category \(c\) is defined as the proportion of consensus-reference pairs in category \(c\) that the source assigns to the same category:
\begin{equation}
\mathrm{Recall}^{(s)}_{c}
=
\frac{C^{(s)}_{c,c}}{\sum_{a \in \mathcal{C}} C^{(s)}_{a,c}}.
\end{equation}

We also summarized severe category reversals between Highly Recommended and Low Fit. Severe over-recommendation was defined as a consensus-reference Low Fit pair assigned as Highly Recommended:
\begin{equation}
\mathrm{OverRec}^{(s)}
=
C^{(s)}_{\mathrm{Highly\ Recommended},\mathrm{Low\ Fit}}.
\end{equation}
Severe under-recommendation was defined as a consensus-reference Highly Recommended pair assigned as Low Fit:
\begin{equation}
\mathrm{UnderRec}^{(s)}
=
C^{(s)}_{\mathrm{Low\ Fit},\mathrm{Highly\ Recommended}}.
\end{equation}
The total number of severe high-low reversals was calculated as
\begin{equation}
\mathrm{Reversal}^{(s)}
=
\mathrm{OverRec}^{(s)}+\mathrm{UnderRec}^{(s)}.
\end{equation}

We analyzed clinician screening time in the NCI counterbalanced TrialGPT 2.0 assistance experiment. Screening time was recorded at the patient-trial-pair level and represented clinician decision time for each candidate trial. Patient-context review time and automated TrialGPT 2.0 PDF report-generation time were not included. For each case \(q\), total screening time with TrialGPT 2.0 assistance was calculated as
\begin{equation}
T^{+}_{q}
=
\sum_{t \in \mathcal{T}_q}
\tau^{+}_{q,t},
\end{equation}
where \(\tau^{+}_{q,t}\) denotes the trial-level screening time recorded under TrialGPT 2.0 assistance. Total screening time without TrialGPT 2.0 assistance was calculated as
\begin{equation}
T^{-}_{q}
=
\sum_{t \in \mathcal{T}_q}
\tau^{-}_{q,t},
\end{equation}
where \(\tau^{-}_{q,t}\) denotes the corresponding trial-level screening time without TrialGPT 2.0 assistance.

Mean total screening time was summarized across cases for the assisted and unassisted conditions. Percentage reduction in screening time was calculated as
\begin{equation}
\mathrm{Reduction}
=
\frac{\overline{T^{-}}-\overline{T^{+}}}{\overline{T^{-}}}
\times 100\%.
\end{equation}
Timing results were summarized overall and within each 50-case counterbalanced group. Error bars in Fig.~\ref{fig:retrospective}d show 95\% confidence intervals of the mean, computed by percentile bootstrap over cases with 10{,}000 resamples. Statistical comparisons between assisted and unassisted screening times used two-sided Wilcoxon signed-rank tests at the case level. Total screening time was also compared between the two clinicians using a two-sided Mann-Whitney U test to assess whether the reduction could be explained by systematic clinician-level speed differences.

\subsection*{Prospective evaluation}

We prospectively evaluated TrialGPT 2.0 in the University of Illinois Cancer Center Precision Oncology Tumor Board (POTB) trial-matching workflow. The analysis included POTB-reviewed cases from February through July 2026 for which TrialGPT 2.0 outputs, routine-workflow-identified trials and final POTB clinical trial recommendations were available. One referred case from April was excluded because the patient was deceased before the POTB session and the case was not included in the final April case list.

For each referred case, the POTB materials summarized the clinical context, molecular findings, prior therapies, and the POTB review question. Trial identification proceeded in parallel: clinical reviewers identified candidate trials through the routine POTB workflow, while TrialGPT 2.0 generated candidate trial recommendations using the UIC trial corpus and matching policy. The two candidate-trial lists were subsequently reviewed together, and clinicians determined the final POTB clinical trial recommendation list. Source attribution reflected whether each final recommendation had been identified in the independently generated TrialGPT 2.0 list, the routine-workflow list or both.

The primary prospective endpoint quantified TrialGPT 2.0's contribution to final clinician-selected trial recommendations at both the case and recommendation levels. At the case level, each reviewed case was assigned to one of three mutually exclusive categories: no final trial recommendation, final trial options expanded by TrialGPT 2.0 or final trial options unchanged by TrialGPT 2.0. A case was classified as expanded when the routine workflow identified no final trial recommendation but TrialGPT 2.0 contributed at least one trial that clinicians selected for the final recommendation list. A case was classified as unchanged at this case-level opportunity threshold when the routine workflow had already identified at least one final recommendation, regardless of whether TrialGPT 2.0 contributed additional recommendations. Cases for which neither source contributed a final recommendation were classified as having no final trial recommendation. We also reported, among cases with at least one final recommendation, the proportion for which TrialGPT 2.0 contributed at least one recommendation and the proportion for which TrialGPT 2.0 was the sole source of the final recommendations (Fig~\ref{fig:prospective}a).

At the recommendation level, each final recommendation was attributed to one of three mutually exclusive sources: TrialGPT 2.0 only, both TrialGPT 2.0 and the routine workflow, or the routine workflow only. TrialGPT 2.0-only recommendations were generated by TrialGPT 2.0 and selected by clinicians for the final recommendation list but were not identified through the routine workflow. Recommendations identified independently by both sources were classified as overlapping recommendations. Routine-workflow-only recommendations were identified through the routine workflow but not by TrialGPT 2.0. TrialGPT 2.0-associated recommendations comprised those attributed to TrialGPT 2.0 only or to both sources. Counts and proportions were summarized overall and by POTB month (Fig~\ref{fig:prospective}b).

The secondary prospective endpoint characterized all TrialGPT 2.0-generated candidate trials using UIC clinician labels. Each candidate trial was labeled as Recommended, Eligible but Not Recommended or Ineligible. Recommended was assigned when the candidate trial was selected for the final POTB recommendation list. Eligible but Not Recommended was assigned when the trial was considered potentially eligible but was not prioritized for final recommendation in the prospective workflow. Ineligible was assigned when the trial was considered incompatible with the patient context or trial criteria. We reported the number and proportion of TrialGPT 2.0-generated candidate trials in each category overall and by POTB month (Fig~\ref{fig:prospective}c).

\subsection*{Reproducible benchmarking}

Reproducible benchmarking evaluation included NIH-TrialBench and previously released public patient-trial matching benchmarks. NIH-TrialBench was used to assess clinician-adjudicated recommendation quality and intended target-trial recovery under synthetic benchmark conditions. Public benchmarks were used to assess standardized eligibility-oriented ranking performance and computational efficiency in comparison with the original TrialGPT framework.

For NIH-TrialBench top-ranked review, TrialGPT 2.0 ranked the predefined NIH-TrialBench candidate trial corpus for each vignette. Clinicians reviewed up to ten highest-ranked trials categorized by TrialGPT 2.0 as Highly Recommended and assigned each reviewed patient-trial pair one of three labels: Recommended, Eligible but Not Recommended or Ineligible. This analysis used the same top-ranked review metric definitions as the retrospective top-ranked review, including eligible precision@$K$, recommended precision@$K$ and hit rate@$K$ for \(K \in \{1,3,5,10\}\), with the case set restricted to NIH-TrialBench vignettes. For these metrics, 95\% confidence intervals were computed by percentile bootstrap over vignettes with 10{,}000 resamples (Fig.~\ref{fig:synthetic}b).

NIH-TrialBench also enabled target-trial assessment because each vignette has a prespecified target trial representing the intended match for the synthetic patient. After TrialGPT 2.0 scored the predefined NIH-TrialBench trial corpus, we located the target trial in the system output and recorded its assigned recommendation category. We summarized the number and proportion of target trials categorized as Highly Recommended, Possible Match or Low Fit.

For the target-trial Recall@10 analysis, each evaluated method ranked candidate trials for each NIH-TrialBench vignette. For TrialGPT 2.0, ranks were generated from the predefined NIH-TrialBench candidate trial corpus. For the off-the-shelf LLM baselines, we queried the ChatGPT web interface using GPT-5.5 Thinking~\cite{openai2026gpt55} and the Gemini web interface using Gemini-3.5 Flash with Extended Thinking~\cite{google2026gemini35}. Baseline queries were conducted on May 22--23, 2026. For each vignette, we started a new temporary chat session to minimize the effect of prior conversation history and submitted a standardized instruction asking the model to search the NIH Clinical Center \textit{Search the Studies} registry~\cite{nih_search_studies} and return only the NCT identifiers of the ten recommended trials for the patient vignette. We used Extended Thinking for Gemini-3.5 Flash because the Standard thinking setting did not reliably return valid trial recommendations under this instruction. Returned NCT identifiers were parsed in the order provided by the model. A target trial was counted as recovered if its NCT identifier appeared among the ten returned identifiers (Supplementary Fig.~\ref{fig:supplementary-synthetic-vignettes}).

Target-trial Recall@10 was calculated separately for each contributing NIH Institute or Center and summarized as an unweighted macro-average across Institutes and Centers. Let $\mathcal{G}$ denote the set of contributing NIH
Institutes and Centers, let $Q_g \subseteq Q_{\mathrm{syn}}$ denote the set of NIH-TrialBench vignettes contributed by institute or center $g$, and let $m$ denote an evaluated method. 

Institute- or center-specific target-trial Recall@10 was defined as
\begin{equation}
\mathrm{Recall@10}_{m,g}
=
\frac{1}{|Q_g|}
\sum_{q \in Q_g}
\mathbb{I}\!\left[
    \operatorname{rank}_{m,q}(t_q) \leq 10
\right],
\label{eq:target-trial-recall-group}
\end{equation}
and macro-averaged target-trial Recall@10 was defined as
\begin{equation}
\mathrm{Recall@10}_{m,\mathrm{macro}}
=
\frac{1}{|\mathcal{G}|}
\sum_{g \in \mathcal{G}}
\mathrm{Recall@10}_{m,g}.
\label{eq:target-trial-recall-macro}
\end{equation}
Here, $t_q$ denotes the prespecified target trial for vignette $q$, and $\operatorname{rank}_{m,q}(t_q)$ denotes the position of that target trial in the ranked list produced by method $m$. When the target trial was not returned among the ten recommendations, its rank was treated as greater than 10. The Avg bars in Fig.~5c report $\operatorname{Recall@10}_{m,\mathrm{macro}}$, giving each contributing NIH Institute or Center equal weight regardless of the number of vignettes it contributed.

For public benchmark evaluation, we evaluated TrialGPT 2.0 on the 2016 Special Interest Group on Information Retrieval (SIGIR) patient-trial matching collection and the TREC 2021 and TREC 2022 Clinical Trials tracks~\cite{koopman2016test,soboroff2021overview,roberts2022overview}. These benchmarks provide standardized relevance judgments for patient-trial matching.

The public benchmark analysis evaluated trial-level matching and ranking, not first-stage retrieval. The TrialGPT 2.0 retrieval module was not applied in this stage. For each patient, the trial search space consisted of all trials with available relevance judgments for that patient. TrialGPT 1.0 and TrialGPT 2.0 each scored every labeled patient-trial pair in this search space, and the scored trials were sorted to produce a ranked list for metric calculation. This design avoids treating unlabeled trials as negative examples and isolates the comparison to trial-level assessment and ranking. The evaluated candidate pools included 58 cases and 3{,}835 judged patient-trial pairs for SIGIR, 75 cases and 35{,}832 judged pairs for TREC 2021, and 50 cases and 35{,}394 judged pairs for TREC 2022.

Benchmark relevance judgments were represented as three-level graded scores. In SIGIR, scores of 0, 1 and 2 correspond to irrelevant, potential and eligible trials, respectively. For TREC 2021 and TREC 2022, we used the official graded relevance judgments, in which 0 denotes non-relevant, 1 denotes excluded and 2 denotes eligible. For Precision@10 and nDCG@10, graded relevance scores were used directly. For mean reciprocal rank (MRR) and mean average precision (MAP), graded relevance was converted to binary relevance by treating eligible trials, corresponding to score 2, as relevant. Patient-trial pairs without relevance judgments were excluded from metric calculation.

Let \(q\) denote a patient, \(i\) denote the rank position of a trial in the model-generated ranked list, \(r_{q,i}\) denote the graded relevance score of the trial ranked at position \(i\) for patient \(q\), and \(r^{*}_{q,i}\) denote the graded relevance score at position \(i\) in the ideal ranking for that patient. The maximum relevance score is \(r_{\max}=2\).

Precision@10 is computed as the normalized graded relevance of the top ten ranked trials:
\begin{equation}
\mathrm{Precision@10}(q)
=
\frac{\sum_{i=1}^{10} r_{q,i}}{10 \times r_{\max}}.
\end{equation}

nDCG@10 is computed as
\begin{equation}
\mathrm{nDCG@10}(q)
=
\frac{\mathrm{DCG@10}(q)}{\mathrm{IDCG@10}(q)},
\end{equation}
where
\begin{equation}
\mathrm{DCG@10}(q)
=
\sum_{i=1}^{10}
\frac{r_{q,i}}{\log_2(i+1)}
\end{equation}
and
\begin{equation}
\mathrm{IDCG@10}(q)
=
\sum_{i=1}^{10}
\frac{r^{*}_{q,i}}{\log_2(i+1)}.
\end{equation}
We set nDCG@10 to 0 for cases with \(\mathrm{IDCG@10}(q)=0\).

For MRR and MAP, eligible trials with \(r_{q,i}=2\) were treated as relevant. Reciprocal rank for patient \(q\) is defined as
\begin{equation}
\mathrm{RR}(q)
=
\frac{1}{\min\{i:r_{q,i}=2\}},
\end{equation}
with reciprocal rank set to 0 if no eligible trial appeared in the ranked list. MRR is the mean of \(\mathrm{RR}(q)\) across cases.

Average precision for patient \(q\) is computed as
\begin{equation}
\mathrm{AP}(q)
=
\frac{1}{M_q}
\sum_{i=1}^{N_q}
\mathrm{P@}i(q)\cdot \mathbb{I}(r_{q,i}=2),
\end{equation}
where \(N_q\) is the number of ranked trials for patient \(q\), \(M_q=\sum_{i=1}^{N_q}\mathbb{I}(r_{q,i}=2)\), \(\mathrm{P@}i(q)\) is binary precision among the top \(i\) ranked trials using the same \(r=2\) relevance threshold, and \(\mathbb{I}(r_{q,i}=2)\) indicates whether the trial at rank \(i\) is eligible. We set average precision to 0 for cases with \(M_q=0\). MAP is the mean of \(\mathrm{AP}(q)\) across cases.

Dataset-level metrics were averaged across cases. The overall benchmark average was calculated as the arithmetic mean of the three dataset-level values from SIGIR, TREC 2021 and TREC 2022. For individual benchmark results, approximate 95\% confidence intervals were computed as the mean \(\pm\) 1.96 standard errors across patient queries. For average bars across benchmarks, error bars summarize variation across the three dataset-level means. Relative change for TrialGPT 2.0 compared with TrialGPT 1.0 was calculated as
\begin{equation}
\mathrm{Relative\ change}
=
\frac{\mathrm{Metric}_{\mathrm{TrialGPT\ 2.0}}-\mathrm{Metric}_{\mathrm{TrialGPT\ 1.0}}}
{\mathrm{Metric}_{\mathrm{TrialGPT\ 1.0}}}
\times 100\%.
\end{equation}

Computational efficiency was measured at the patient-trial-pair level using the same judged candidate pools evaluated for ranking. For each scored pair, we recorded latency and model token usage. For TrialGPT 1.0, latency and token counts were summed across the criterion-level matching and aggregation calls for the pair. For TrialGPT 2.0, latency and token counts were recorded from the corresponding trial-level scoring call. Processing speed was defined as the inverse of the mean per-pair latency and reported as patient-trial pairs scored per second. Input tokens included all tokens submitted to the backbone model, and output tokens included all tokens generated by the model. Average speed, input tokens and output tokens were calculated across patient-trial pairs within each benchmark.

The main public benchmark comparison used GPT-4.1~\cite{openai2025gpt41} as the backbone model for both TrialGPT 1.0 and TrialGPT 2.0. Additional analyses with alternative backbone models are reported in Supplementary Fig.~\ref{fig:supplementary-backbone}.

\subsection*{Real-world deployment evaluation}

To facilitate structured clinician review, the TrialGPT 2.0 interface accepts patient context, initiates matching against the predefined local trial corpus and displays the resulting matched-trial list. The interface uses the same TrialGPT 2.0 backend, matching policy and trial-corpus versions as the corresponding evaluation setting. Interface-level filtering, sorting and PDF export are applied after TrialGPT 2.0 generates the matched-trial output and do not change the model-generated recommendation category, rank or structured explanation used for evaluation.

In participating deployment settings, local trial search spaces are refreshed weekly to reflect changes in trial availability, recruitment status, cohort or arm availability and site-specific trial lists. Each recommendation run uses the local trial-corpus snapshot available at the time of matching. For NIH-TrialBench, retrospective clinical-note cohorts, public benchmark analyses and runtime experiments, the trial-corpus or benchmark candidate set was fixed before inference and recorded, so later weekly updates did not alter the reported results. For the prospective POTB evaluation, TrialGPT 2.0 used the current UIC trial-corpus snapshot available at the time of each case review, reflecting the weekly update process used in the live workflow; the snapshot used for each prospective run was recorded for auditability.

We measured recommendation-generation time in the interface deployment environment. Timing started when the user initiated matching after entering the patient context and ended when the matched-trial list was displayed in the interface. The measurement included retrieval when applicable, trial loading, backend LLM-based assessment, score aggregation and ranking, output parsing, result writing and score-sorted PDF generation. It excluded patient-context entry time, manual clinician review and any post-generation human actions. For the runtime experiment, we sampled five cases from each evaluation source and ran each case three times using the corresponding site-specific trial corpus and matching configuration, in the interface runtime environment using 128 worker processes. The three repeated runs were averaged to obtain a case-level runtime estimate, and runtime was summarized across cases within each source as the mean and standard deviation (Supplementary Fig.~\ref{fig:supplementary-time-cost}).

Clinical reviewer acceptance was assessed using an 8-item questionnaire after reviewers used or reviewed TrialGPT 2.0 outputs in clinical trial matching workflows. The questionnaire assessed usability, workflow fit, perceived efficiency, clarity, decision confidence, usefulness, content safety and overall satisfaction using a 5-point Likert scale, where 1 indicated strongly disagree and 5 indicated strongly agree. We summarized item-level response distributions and favorable responses, defined as ratings of 4 or 5 (Supplementary Fig.~\ref{fig:supplementary-reviewer-acceptance-responses}).

\subsection*{Study approval and ethics oversight}

The study was conducted in accordance with applicable institutional policies and regulatory requirements. Retrospective and prospective clinical-note cohorts were analyzed using de-identified, coded or anonymized data under applicable institutional approvals, waivers, data-use agreements or non-human-subjects research determinations at the contributing sites. TrialGPT 2.0 outputs were used to support clinician review; clinicians retained responsibility for final trial recommendations, referral, screening and enrollment decisions. Results are reported only in aggregate, and no directly identifiable participant information is included in the manuscript.

The National Cancer Institute retrospective cohort, NIH Office of Patient Recruitment cohort, and Cholangiocarcinoma Foundation cohort were analyzed using only de-identified or anonymized data and were determined to constitute Not Human Subjects Research (NHSR) under applicable federal regulations~\cite{nih_nhsr}. Accordingly, informed consent was not required. No biospecimens were accessed, no participants were contacted for research purposes, and no additional clinical data were collected beyond information already available from routine referral, recruitment, or trial-matching workflows.

The University of Pittsburgh Medical Center cohort was analyzed under approval of the UPMC Institutional Review Board (STUDY20070273), with waivers of informed consent as applicable.

The University of Illinois Cancer Center retrospective and prospective Precision Oncology Tumor Board evaluations were conducted under UIC protocol STUDY2025-0124, with a waiver of documentation of informed consent and a waiver or alteration of HIPAA authorization.

NIH-TrialBench used clinician-authored synthetic vignettes constructed from target trials and did not use private patient records.

\bibliography{reference}

\section*{Acknowledgements}
This research was supported by the Intramural Research Program of the National Institutes of Health (NIH). The contributions of the NIH author(s) are considered Works of the United States Government. This research was also partially supported by the NIH Pathway to Independence Award K99LM014903 (Q.J.). The findings and conclusions presented in this paper are those of the author(s) and do not necessarily reflect the views of the NIH or the U.S. Department of Health and Human Services.

We thank Juhi Gor and Aseem Aseem at the University of Illinois Cancer Center for data curation support for the retrospective and prospective Precision Oncology Tumor Board evaluations. We thank Zhizheng Wang at the National Library of Medicine, National Institutes of Health, for his valuable feedback on this work. We also thank the clinical reviewers, trial navigators, coordinators and recruitment staff at the participating sites for their contributions to clinical trial review workflows and feedback on TrialGPT 2.0 outputs.

\section*{NIH-TrialBench Consortium}

\textbf{National Cancer Institute (NCI)}: Stacey Doran, MD; Brooke Kania, DO; Charalampos S. Floudas, MD, DMSc, MS; Yoonhee Choi, MD; Aanika Warner, MD, MPH; Padma Rajagopal, MD, MPH, MSc; and Jung-Min Lee, MD.

\textbf{National Human Genome Research Institute (NHGRI)}: Benjamin Solomon, MD; Christopher Ours, MD, MHS; and Precilla D'Souza, DNP, MSN, CRNP.

\textbf{National Eye Institute (NEI)}: Mélanie Hébert, MD, MSc; Asmita Indurkar, MBBS, MS; Emily Chew, MD; and Tiarnán Keenan, MD, PhD.

\textbf{National Heart, Lung, and Blood Institute (NHLBI)}: Emma Groarke, MD; and Gonca Ozcan, MD.

\textbf{National Institute of Mental Health (NIMH)}: Mark Kvarta, MD, PhD.

\textbf{National Institute of Dental and Craniofacial Research (NIDCR)}: Vivian Macdonald-Szymczuk, MD; Iris Hartley, MD; and Konstantinia Almpani, DDS, MSc.

\textbf{National Institute of Allergy and Infectious Diseases (NIAID)}: Elise O'Connell, MD; and Daniel Rogan, MD.

\textbf{National Institute of Environmental Health Sciences (NIEHS)}: Stavros Garantziotis, MD.

\textbf{National Institute on Drug Abuse (NIDA)}: Shiva Pandey, PA-C.

\textbf{Eunice Kennedy Shriver National Institute of Child Health and Human Development (NICHD)}: Christina Tatsi, MD, MHSc, PhD.

\section*{Author contributions}
Z.L. conceived the study. Y.F., Q.J., C.S.F.  and Z.L. designed the TrialGPT 2.0 system evaluation. Y.F. developed the TrialGPT 2.0 system. S.T. implemented the TrialGPT 2.0 interface. L.H. coordinated the collection of synthetic vignettes from NIH Institutes and Centers. L.H. and M.G. provided input on clinical recruitment workflows. Y.F. performed the primary analyses, generated figures, and drafted the manuscript with input from Q.J., Z.W., J.S., and Z.L. N.N. and R.H.-T.N. led the UIC retrospective and prospective Precision Oncology Tumor Board evaluations. C.S.F. contributed to the design of the NCI retrospective evaluation and provided the 100 de-identified clinical vignettes. K.M. and C.M.M. performed the timed exhaustive-review annotations, and J.L.G. contributed clinical trial recruitment-workflow expertise. A.N., J.V., M. Bachini, L.R.-N. and K.R. contributed the CCF cohort and clinical review. N.C., M. Barnes and M.M. contributed the OPR cohort and recruitment-workflow expertise. D.G., C.E.G., H.S., and M. Burczynski contributed the UPMC cohort and clinical review. The NIH-TrialBench Consortium authored target-trial-informed synthetic vignettes and contributed clinical domain expertise for the NIH-TrialBench benchmark. Z.L. supervised the study. All authors interpreted the results, revised the manuscript, and approved the final version. 

\section*{Competing interests}
The authors declare no competing interests.

\setcounter{figure}{0}
\clearpage
\captionsetup[figure]{name=Supplementary Fig.}
\captionsetup[table]{name=Supplementary Table,labelsep=period}
\setcounter{table}{0}
\section*{Supplementary Information}

\textbf{Supplementary Notes}
\begin{itemize}
  \item \hyperref[suppnote:factor]{Supplementary Note 1: Trial-fit factor coverage across matching policies}
  \item \hyperref[suppnote:score-validity]{Supplementary Note 2: Fit and confidence scores respond to clinical mismatch and missing information}
  \item \hyperref[suppnote:clinician-comments]{Supplementary Note 3: Clinician-comment analysis of non-recommended outputs}
  \item \hyperref[suppnote:public-benchmark]{Supplementary Note 4: Public benchmark comparison of TrialGPT 1.0 and TrialGPT 2.0}
  \item \hyperref[suppnote:interface-time]{Supplementary Note 5: Recommendation-generation time in the TrialGPT 2.0 interface}
  \item \hyperref[suppnote:reviewer-acceptance]{Supplementary Note 6: Clinical reviewer acceptance of TrialGPT 2.0}
\end{itemize}

\textbf{Supplementary Tables}
\begin{itemize}
  \item \hyperref[tab:factor_coverage]{Supplementary Table 1: Depth of trial-fit factor assessment by matching policy}
  \item \hyperref[tab:clinical_reviewer_acceptance]{Supplementary Table 2: Clinical reviewer acceptance questionnaire}
\end{itemize}

\textbf{Supplementary Figures}
\begin{itemize}
  \item \hyperref[fig:validity]{Supplementary Fig. 1: TrialGPT 2.0 score behavior under clinical mismatch and missing-information perturbations}
  \item \hyperref[fig:supplementary-errors]{Supplementary Fig. 2: Clinician-comment analysis of discordant TrialGPT 2.0 Highly Recommended patient-trial pairs}
  \item \hyperref[fig:supplementary-backbone]{Supplementary Fig. 3: Public benchmark comparison of TrialGPT 2.0 and TrialGPT 1.0 across LLM backbones}
  \item \hyperref[fig:interface]{Supplementary Fig. 4: TrialGPT 2.0 web-based interface for trial recommendation review}
  \item \hyperref[fig:supplementary-time-cost]{Supplementary Fig. 5: TrialGPT 2.0 recommendation-generation runtime across evaluation sources}
  \item \hyperref[fig:supplementary-reviewer-acceptance-responses]{Supplementary Fig. 6: Clinical reviewer acceptance responses for TrialGPT 2.0}
  \item \hyperref[fig:supplementary-synthetic-vignettes]{Supplementary Fig. 7: NIH-TrialBench target-trial recovery}
\end{itemize}
\clearpage

\phantomsection\label{suppnote:factor}
\subsection*{Supplementary Note 1: Trial-fit factor coverage across matching policies}

The trial-fit factors shown in Fig.~\ref{fig:overview} represent factors that TrialGPT 2.0 can assess rather than a fixed checklist applied identically in every workflow. Each site- or evaluation-specific matching policy determines which factors are represented and the depth with which they are assessed. Factor coverage was categorized as follows:
\begin{itemize}[leftmargin=1.2em, itemsep=-0.2em, topsep=-0.2em]
    \item \textbf{Dedicated}: the matching policy contains specific instructions for assessing the factor.
    \item \textbf{Partial}: the factor is considered only in certain circumstances, through the raw trial eligibility criteria or through instructions related to another factor, rather than through separate factor-specific instructions.
    \item \textbf{Triage}: the factor is used for high-level screening based on brief referral information rather than comprehensive assessment.
\end{itemize}

These categories describe how each factor is represented in the matching policy. All matching policies receive the patient context and raw trial eligibility criteria; therefore, a factor categorized as Partial may still influence the assessment when relevant information is available. Therapeutic benefit was not categorized separately because it was treated as a composite clinical judgment reflected in the overall fit assessment and rationale rather than as a standalone factor-specific component.

\paragraph{Diagnosis or histology.}
All matching policies assess diagnosis or disease subtype. Histology is considered when specified in the raw trial eligibility criteria. UIC, CCF, UPMC, NCI, and NIH-TrialBench contain dedicated diagnosis-matching instructions, whereas OPR uses diagnosis or phenotype primarily for referral triage.

\paragraph{Disease status.}
UIC and CCF assess disease and stage fit through dedicated instructions, including factors such as advanced or metastatic disease, measurable disease and treatment-naive versus previously treated settings. UPMC explicitly considers selected disease-status characteristics, including measurable disease, recurrence or second-primary status and prior curative treatment, but does not contain a comprehensive disease-status component. NCI, OPR and NIH-TrialBench consider disease fit through the patient context, trial criteria and broader relevance assessment, without a dedicated component covering metastatic, recurrent, stage or no-evidence-of-disease status.

\paragraph{Biomarker or pathway match.}
UIC, CCF, NCI and NIH-TrialBench contain explicit instructions for assessing required biomarkers, biomarker-defined disease and biomarker-specific cohorts. Pathway relevance is also emphasized in UIC and CCF but is not separately represented in the NCI and NIH-TrialBench policy. UPMC considers biomarkers and genomic categories primarily in relation to cohort or arm assignment. OPR treats the absence of a required biomarker as an important screening concern but does not emphasize broader pathway matching.

\paragraph{Line of therapy.}
UIC and CCF contain dedicated line-of-therapy components that compare the patient's prior treatment-line count with trial-level or cohort-level requirements. UPMC, NCI, NIH-TrialBench and OPR consider line of therapy when relevant information is present in the patient context or raw trial criteria, but do not contain dedicated line-counting components.

\paragraph{Prior treatment.}
UIC, CCF and UPMC explicitly assess prior systemic therapy, treatment timing, washout requirements and prohibited or concurrent therapies. NCI, OPR and NIH-TrialBench focus primarily on prohibited prior therapies or prior-treatment information directly relevant to the trial criteria.

\paragraph{Cohort or arm applicability.}
UIC, CCF, UPMC, NCI and NIH-TrialBench contain explicit logic for evaluating umbrella or master protocols and determining whether the patient fits at least one relevant cohort or arm. OPR considers alignment with the referral goal and broad cohort preferences but does not perform comprehensive multi-arm assignment.

\paragraph{Inclusion and exclusion criteria.}
UIC, CCF, UPMC, NCI and NIH-TrialBench use the raw trial inclusion and exclusion criteria as the primary basis for formal eligibility assessment. OPR applies these criteria in a triage-oriented manner, emphasizing clear hard exclusions, demographic mismatches, numeric-threshold violations and alignment with the referral goal.

\begin{table}[!t]
\centering
\caption{\textbf{Depth of trial-fit factor assessment by matching policy.} Symbols indicate whether each factor is addressed through specific policy instructions, considered partially, or applied at a triage level.}
\label{tab:factor_coverage}
\small
\setlength{\tabcolsep}{3pt}
\renewcommand{\arraystretch}{1.35}

\renewcommand{\tabularxcolumn}[1]{m{#1}}

\newcommand{\symDed}{{\large $\bullet$}}
\newcommand{\symPar}{{\large $\circ$}}
\newcommand{\symTri}{{\large $\triangle$}}

\begingroup
\scriptsize
\arrayrulecolor{gray!55}
\setlength{\arrayrulewidth}{0.4pt}

\begin{tabularx}{0.98\linewidth}{|>{\raggedright\arraybackslash}m{0.27\linewidth}|>{\centering\arraybackslash}X|>{\centering\arraybackslash}X|>{\centering\arraybackslash}X|>{\centering\arraybackslash}X|>{\centering\arraybackslash}X|>{\centering\arraybackslash}X|}
\hline
\rowcolor{gray!10}
\textbf{Trial-fit factor} & 
\textbf{OPR} & 
\textbf{CCF} & 
\textbf{UIC} & 
\textbf{UPMC} & 
\textbf{NCI} & 
\textbf{NIH-TrialBench} \\
\hline
Diagnosis or histology & \symTri & \symDed & \symDed & \symDed & \symDed & \symDed \\
\hline
Disease status & \symPar & \symDed & \symDed & \symPar & \symPar & \symPar \\
\hline
Biomarker or pathway match & \symPar & \symDed & \symDed & \symPar & \symDed & \symDed \\
\hline
Line of therapy & \symPar & \symDed & \symDed & \symPar & \symPar & \symPar \\
\hline
Prior treatment & \symPar & \symDed & \symDed & \symDed & \symPar & \symPar \\
\hline
Cohort or arm applicability & \symPar & \symDed & \symDed & \symDed & \symDed & \symDed \\
\hline
Inclusion and exclusion criteria & \symTri & \symDed & \symDed & \symDed & \symDed & \symDed \\
\hline
\rowcolor{gray!10}
\multicolumn{7}{|l|}{\textbf{Key:} \quad \symDed = Dedicated \quad\quad \symPar = Partial \quad\quad \symTri = Triage} \\
\hline
\end{tabularx}
\endgroup
\end{table}

\phantomsection\label{suppnote:score-validity}
\subsection*{Supplementary Note 2: Fit and confidence scores respond to clinical mismatch and missing information}

\begin{figure}[!th]
  \centering
  \includegraphics[width=\textwidth,height=0.62\textheight,keepaspectratio]{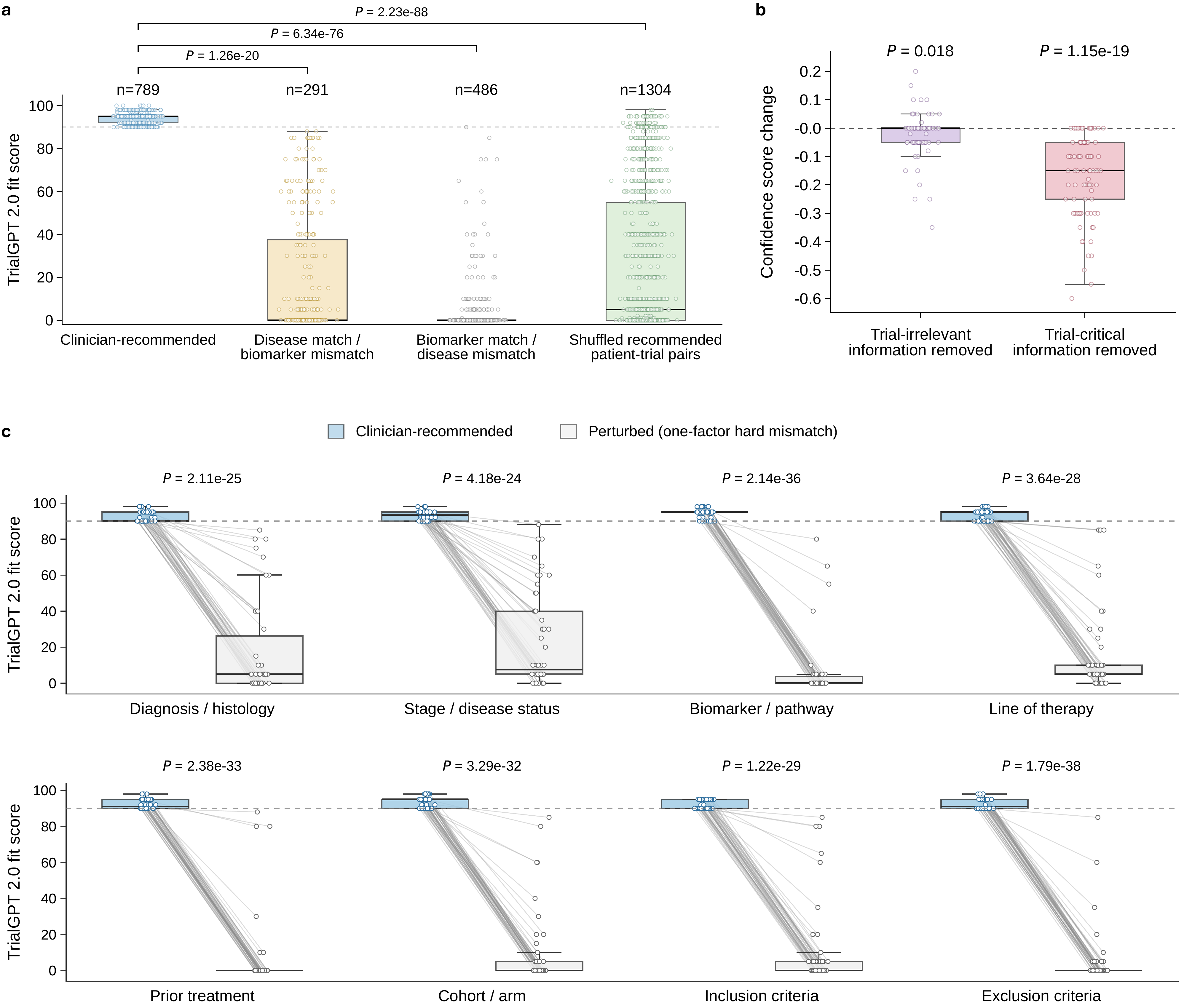}
  \caption{\textbf{TrialGPT 2.0 score behavior under clinical mismatch and missing-information perturbations.}
  \textbf{a,} Fit-score discrimination among clinician-recommended patient-trial pairs, disease-matched pairs with biomarker mismatch, biomarker-matched pairs with disease mismatch and shuffled clinician-recommended patient-trial pairs. $P$ values were obtained using two-sided Welch's $t$-tests comparing patient-level mean fit scores for clinician-recommended trials versus each control condition.
  \textbf{b,} Confidence-score change after removing trial-irrelevant or trial-critical information from patient summaries. $P$ values were obtained using two-sided paired $t$-tests comparing the original confidence score with the confidence score after information removal for the same patient-trial pair.
  \textbf{c,} Fit-score sensitivity to counterfactual one-factor hard mismatches in diagnosis or histology, stage or disease status, biomarker or pathway, line of therapy, prior treatment, cohort or arm applicability, inclusion criteria and exclusion criteria. Gray lines connect paired original and perturbed scores for the same patient-trial anchor. $P$ values were obtained using two-sided paired $t$-tests comparing the original fit score with the perturbed fit score for the same patient-trial anchor.  Dashed horizontal lines indicate the Highly Recommended threshold in \textbf{a} and \textbf{c} and no confidence-score change in \textbf{b}. Boxes show the interquartile range, center lines show medians and whiskers extend to the most extreme values within 1.5$\times$ the interquartile range. Points indicate individual patient-trial pairs in \textbf{a}, paired perturbation examples in \textbf{b} and paired original or perturbed observations in \textbf{c}.
  }
  \label{fig:validity}
\end{figure}

We performed complementary analyses to evaluate whether the TrialGPT 2.0 fit and confidence scores behave as intended. We first tested whether the fit score distinguishes patient-specific compatibility from partial or nonspecific relevance. We then tested whether it responds to isolated incompatibilities in individual trial-fit factors. Finally, we examined whether the confidence score responds selectively to the removal of information needed to assess a candidate trial.

\subsubsection*{Fit-score discrimination using partial-match and patient-shuffled controls}

For the fit-score comparison, we used existing TrialGPT 2.0 matching outputs from NIH-TrialBench and the NCI, UIC and CCF retrospective cohorts. These evaluation sources were selected because their matching policies contained specific instructions for both diagnosis or disease matching and biomarker matching, the two dimensions used to construct the partial-match controls (Supplementary Note~1 and Supplementary Table~\ref{tab:factor_coverage}). Clinician-recommended pairs were defined as patient-trial pairs that TrialGPT 2.0 categorized as Highly Recommended and clinicians labeled as Recommended. We compared these pairs with three control groups designed to preserve apparent relevance while disrupting the patient-specific match (Supplementary Fig.~\ref{fig:validity}a).

The first two groups were partial-match controls selected from each patient's scored TrialGPT 2.0 output. Disease-matched/biomarker-mismatched controls were patient-trial pairs in which the trial appeared relevant to the patient's disease but required a different or undocumented biomarker or pathway. Biomarker-matched/disease-mismatched controls were pairs in which the patient and trial shared a relevant biomarker or pathway, but the trial disease context did not match the patient.

Candidate partial-match pairs were reviewed using GPT-5.4~\cite{openai2026gpt54}, with the patient summary, trial criteria and TrialGPT 2.0 explanation provided as inputs. The TrialGPT 2.0 explanation was included to clarify the assessed rationale and support review of ambiguous multi-arm or basket-trial contexts. A candidate was retained only when the assigned mismatch category was confirmed and no plausible applicable cohort or arm was identified.

The third group consisted of patient-shuffled controls. Starting from a clinician-recommended pair comprising patient A and trial X, we identified other patients for whom trial X also appeared in the scored TrialGPT 2.0 output. Each resulting pair between trial X and a different patient was retained as a shuffled control, using the fit score already assigned by TrialGPT 2.0 to that patient-trial pair.

Fit scores were aggregated to patient-level means, and clinician-recommended pairs were compared with each control condition using two-sided Welch's $t$-tests. Clinician-recommended pairs received consistently high fit scores, whereas both partial-match controls scored substantially lower. These results indicate that the fit score does not rely on disease or biomarker similarity alone. Patient-shuffled controls also scored lower, consistent with the fit score reflecting the specific patient-trial pairing rather than trial-level attractiveness (Supplementary Fig.~\ref{fig:validity}a).

\subsubsection*{Fit-score sensitivity to counterfactual hard mismatches}

We next tested whether the fit score responds to individual trial-fit factors (Fig.~\ref{fig:overview}b). We used clinician-recommended patient-trial pairs from NIH-TrialBench and the retrospective cohorts as positive anchors. For each anchor, GPT-5.4 generated a minimally edited patient profile that disrupted one prespecified trial-fit factor while keeping the trial text fixed and preserving the remainder of the patient profile whenever possible. For example, a biomarker perturbation changed a required positive biomarker to negative or mutually exclusive, whereas a stage perturbation changed the disease status to one incompatible with the trial requirement (Supplementary Fig.~\ref{fig:validity}c).

Generated perturbations were retained only after quality-control review. A retained perturbation had to satisfy five criteria: the original patient-trial pair was clinician-recommended; the targeted factor was relevant to the candidate trial, as indicated by the raw trial eligibility criteria or the corresponding matching policy; the edited patient profile was internally consistent; the edit targeted the specified factor without introducing unrelated changes; and the edited patient was clearly incompatible with the trial for that factor. Multi-arm and basket trials were reviewed carefully, and perturbations were excluded when another arm or cohort remained plausibly applicable.

We did not perturb therapeutic relevance or benefit separately because it represents a composite clinical judgment rather than a single objective trial-fit variable that can be minimally changed without altering multiple clinical facts.

TrialGPT 2.0 was rerun on each original and counterfactual patient-trial pair using the same trial and site-specific prompt context. For Supplementary Fig.~\ref{fig:validity}c, we used a balanced set of 50 quality-controlled perturbations for each factor group: diagnosis or histology, stage or disease status, biomarker or pathway, line of therapy, prior treatment, cohort or arm applicability, inclusion criteria and exclusion criteria. Each counterfactual fit score was compared with the original fit score for the corresponding clinician-recommended anchor pair using a two-sided paired $t$-test.

Single-factor hard mismatches produced marked reductions in fit score across all evaluated domains. These results show that the fit score responds to clinically important incompatibilities even when the trial and the remainder of the patient profile are held constant (Supplementary Fig.~\ref{fig:validity}c).

\subsubsection*{Confidence-score sensitivity to missing information}

The confidence score addresses a different property: whether the available patient information is sufficient to support the fit judgment. To evaluate this behavior, we used 100 original patient-trial anchor pairs from NIH-TrialBench and the retrospective cohorts. For each pair, GPT-5.4 generated two minimally edited patient profiles while keeping the candidate trial fixed: one with trial-irrelevant information removed and one with trial-critical information removed. This yielded 100 trial-irrelevant information-removal pairs and 100 trial-critical information-removal pairs (Supplementary Fig.~\ref{fig:validity}b).

Trial-irrelevant information comprised patient details that were not expected to affect assessment of the candidate trial, such as unrelated medical history, non-decisive comorbidities or details not referenced by the trial criteria. Trial-critical information comprised evidence needed to assess eligibility or clinical fit, such as biomarker status, disease stage or status, line of therapy, prior treatment exposure, ECOG performance status, organ function or central nervous system involvement.

Generated perturbations were retained only after quality-control review. The removed information had to be present in the original patient profile, the edited profile had to remain internally consistent, and the edit had to remove information rather than replace it with an incompatible value. This distinction separated missing-information perturbations from the counterfactual hard mismatches described above: missing or obscured evidence was expected to affect confidence, whereas direct incompatibility was expected to affect fit.

TrialGPT 2.0 was rerun on the same candidate trial using the perturbed patient context and the same site-specific prompt context. For each pair, we calculated the confidence-score change relative to the original assessment. Changes were summarized separately for trial-irrelevant and trial-critical information removal, and the confidence score under each removal condition was compared with the corresponding original score using a two-sided paired $t$-test.

Removing trial-irrelevant information produced only small confidence changes, whereas removing trial-critical information led to larger reductions. These results support the intended role of the confidence score in reflecting whether key information needed for patient-trial assessment is available (Supplementary Fig.~\ref{fig:validity}b).

\phantomsection\label{suppnote:clinician-comments}
\subsection*{Supplementary Note 3: Clinician-comment analysis of non-recommended outputs}

\begin{figure}[!ht]
  \centering
  \includegraphics[width=\textwidth]{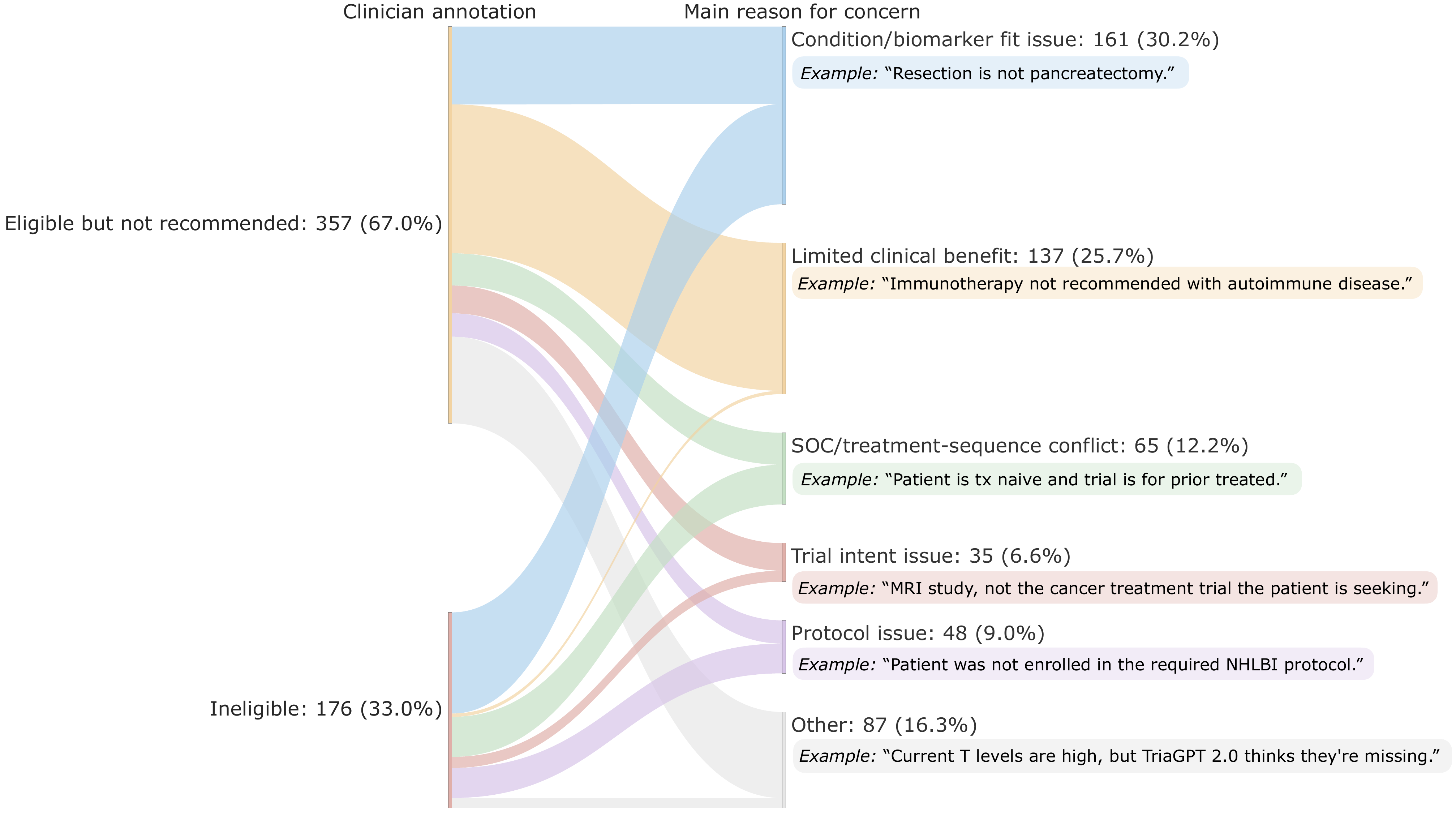}
  \caption{\textbf{Clinician-comment analysis of discordant TrialGPT 2.0 Highly Recommended patient-trial pairs.}
  The alluvial plot summarizes patient-trial pairs assigned to the Highly Recommended category by TrialGPT 2.0 but labeled by clinicians as Eligible but Not Recommended or Ineligible. The left side shows clinician annotations, and the right side shows the main reason for concern identified from clinician comments, including condition or biomarker fit issues, limited clinical benefit, standard-of-care (SOC) or treatment-sequence conflicts, trial-intent issues, protocol issues and other concerns. Counts and percentages are shown for each annotation and concern category, with representative clinician comments displayed as examples.
  }
  \label{fig:supplementary-errors}
\end{figure}

To understand why some top-ranked TrialGPT 2.0 recommendations were not judged clinically appropriate, we analyzed free-text clinician comments attached to patient-trial pairs that TrialGPT 2.0 categorized as Highly Recommended but clinicians labeled as Eligible but Not Recommended or Ineligible.

Each comment was assigned one primary reason for concern according to the main issue expressed by the clinician. Categories comprised condition or biomarker fit issue, limited clinical benefit, standard-of-care or treatment-sequence conflict, trial-intent issue, protocol issue and other. Condition or biomarker fit issues included mismatches in disease, histology, molecular subtype or required biomarker. Limited clinical benefit included cases in which a trial was technically possible but was judged to offer low value because of expected benefit, toxicity, patient goals or better available options. Standard-of-care or treatment-sequence conflicts included concerns related to treatment timing, prior-therapy requirements or line-of-therapy logic. Trial-intent issues included studies whose purpose did not match the patient's clinical need, such as imaging, screening, registry, prevention or supportive-care studies when therapeutic treatment was sought. Protocol issues included specific eligibility criteria, required protocol participation, cohort rules, enrollment requirements or trial-status concerns. Comments that did not fit these categories or were too nonspecific to assign confidently were labeled as other.

Category assignment was performed at the individual-comment level. When a comment contained multiple concerns, we assigned the category corresponding to the dominant reason that the clinician rejected or deprioritized the trial. Assignments were manually reviewed for consistency, with particular attention to overlapping themes such as limited benefit versus trial intent and biomarker mismatch versus cohort or protocol mismatch. Comments that did not contain a substantive clinical concern were not assigned to a reason-for-concern category.

Reason-for-concern categories were summarized overall and stratified by clinician annotation, comparing Eligible but Not Recommended with Ineligible cases. Clinician comments identified both patient-trial compatibility concerns, including condition, biomarker and protocol mismatches, and clinical-priority concerns, including limited expected benefit, treatment-sequence conflicts and mismatch between trial intent and the patient's care goal. The presence of many Eligible but Not Recommended annotations indicates that discordant outputs were not limited to binary eligibility failures, but frequently reflected clinical judgment about trial appropriateness and priority (Supplementary Fig.~\ref{fig:supplementary-errors}).

\phantomsection\label{suppnote:public-benchmark}
\subsection*{Supplementary Note 4: Public benchmark comparison of TrialGPT 1.0 and TrialGPT 2.0}

\begin{figure}[H]
  \centering
  \includegraphics[width=\textwidth]{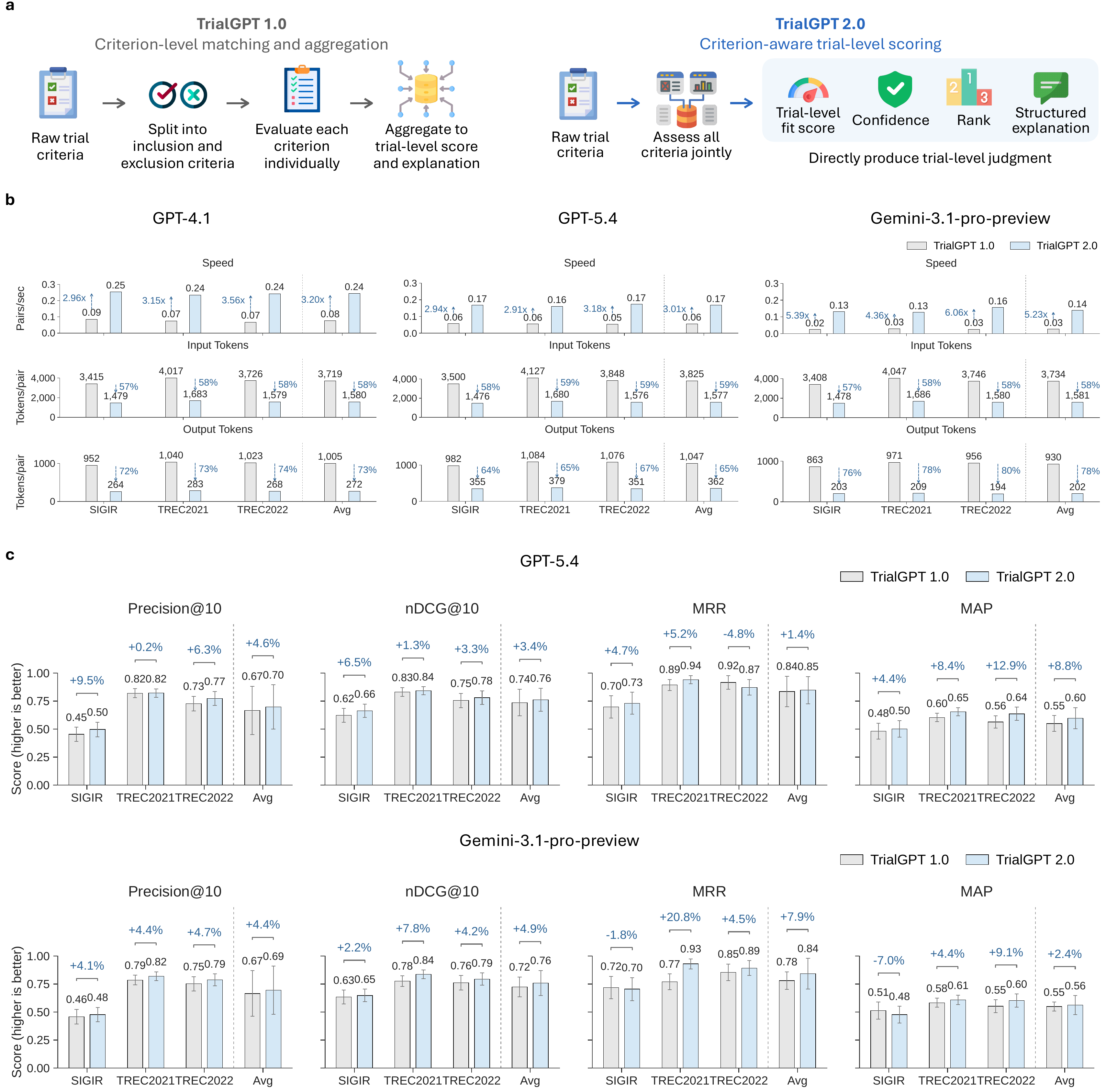}
  \caption{\textbf{Public benchmark comparison of TrialGPT 2.0 and TrialGPT 1.0 across LLM backbones.}
  \textbf{a,} TrialGPT 2.0 replaces the TrialGPT 1.0 criterion-level aggregation workflow with criterion-aware trial-level scoring.
  \textbf{b,} Computational efficiency across SIGIR, TREC2021 and TREC2022 using GPT-4.1~\cite{openai2025gpt41}, GPT-5.4~\cite{openai2026gpt54} and Gemini-3.1-pro-preview~\cite{googleai2026gemini31propreview}, measured by matching speed and token use per patient-trial pair. Percentages indicate relative changes for TrialGPT 2.0 versus TrialGPT 1.0.
  \textbf{c,} Eligibility-ranking performance across public benchmarks and LLM backbones, evaluated using Precision@10, normalized discounted cumulative gain at 10 (nDCG@10), mean reciprocal rank (MRR) and mean average precision (MAP). Bars denote mean performance. For individual benchmarks, error bars show approximate 95\% confidence intervals (mean $\pm$ 1.96 $\times$ standard error across patient queries); for Avg bars, error bars summarize variation across the three benchmark-level means. Percentage annotations indicate the relative change of TrialGPT 2.0 compared with TrialGPT 1.0.
  }
  \label{fig:supplementary-backbone}
\end{figure}

We evaluate TrialGPT 2.0 on established public patient-trial matching benchmarks, including the 2016 Special Interest Group on Information Retrieval (SIGIR) test collection~\cite{koopman2016test} and the 2021 and 2022 Clinical Trials tracks of the Text REtrieval Conference (TREC)~\cite{soboroff2021overview,roberts2022overview}. 

We compare TrialGPT 2.0 with the original TrialGPT framework~\cite{jin2024matching}, referred to here as TrialGPT 1.0. TrialGPT 1.0 performs criterion-level matching: it decomposes raw eligibility criteria into inclusion and exclusion criteria, evaluates each criterion against the patient record and aggregates criterion-level outputs into a trial-level score and explanation. TrialGPT 2.0 retains criterion-aware assessment but generates trial-level judgments directly, reviewing trial criteria jointly and returning a fit score, confidence estimate, rank and structured explanation for each trial (Supplementary Fig.~\ref{fig:supplementary-backbone}a).

We assess ranking performance using Precision@10~\cite{schutze2008introduction}, nDCG@10~\cite{jarvelin2002cumulated}, mean reciprocal rank (MRR)~\cite{voorhees1999trec} and mean average precision (MAP)~\cite{schutze2008introduction}, standard information-retrieval metrics that capture complementary aspects of ranked retrieval quality, including top-ranked precision, rank-sensitive relevance, early retrieval of relevant trials and ranking quality across relevant trials. Across SIGIR, TREC 2021 and TREC 2022, TrialGPT 2.0 improves all evaluated ranking metrics relative to TrialGPT 1.0. Averaged across benchmarks, TrialGPT 2.0 increases Precision@10 by 4.5\%, nDCG@10 by 5.7\%, MRR by 9.8\% and MAP by 10.2\% (Supplementary Fig.~\ref{fig:supplementary-backbone}c). These gains indicate that joint trial-level assessment improves benchmark ranking quality, a prerequisite for using TrialGPT 2.0 to prioritize candidate trials for clinician review.

TrialGPT 2.0 also reduces inference burden. Averaged across benchmarks, processing speed increases by 3.20-fold, while input and output tokens per patient-trial pair decrease by 58\% and 73\%, respectively (Supplementary Fig.~\ref{fig:supplementary-backbone}b). This efficiency gain is consistent with the architectural change from criterion-level aggregation to joint trial-level assessment: criterion-level matching requires repeated assessment of individual eligibility criteria and longer intermediate outputs, whereas TrialGPT 2.0 produces a compact structured assessment for each patient-trial pair. In the web-based implementation, candidate-trial assessments run in parallel using multiprocessing, allowing the interface to display ranked matching results for one case in approximately 22 seconds on average.

Additional analyses with GPT-5.4~\cite{openai2026gpt54} and Gemini-3.1-pro-preview~\cite{googleai2026gemini31propreview} show a similar overall pattern of improved efficiency and favorable average ranking performance with TrialGPT 2.0 (Supplementary Fig.~\ref{fig:supplementary-backbone}). We use GPT-4.1 in the web-based implementation because it provides a favorable trade-off between ranking performance and computational efficiency.

\phantomsection\label{suppnote:interface-time}
\subsection*{Supplementary Note 5: Recommendation-generation time in the TrialGPT 2.0 interface}

\begin{figure}[!ht]
  \centering
  \includegraphics[width=\textwidth]{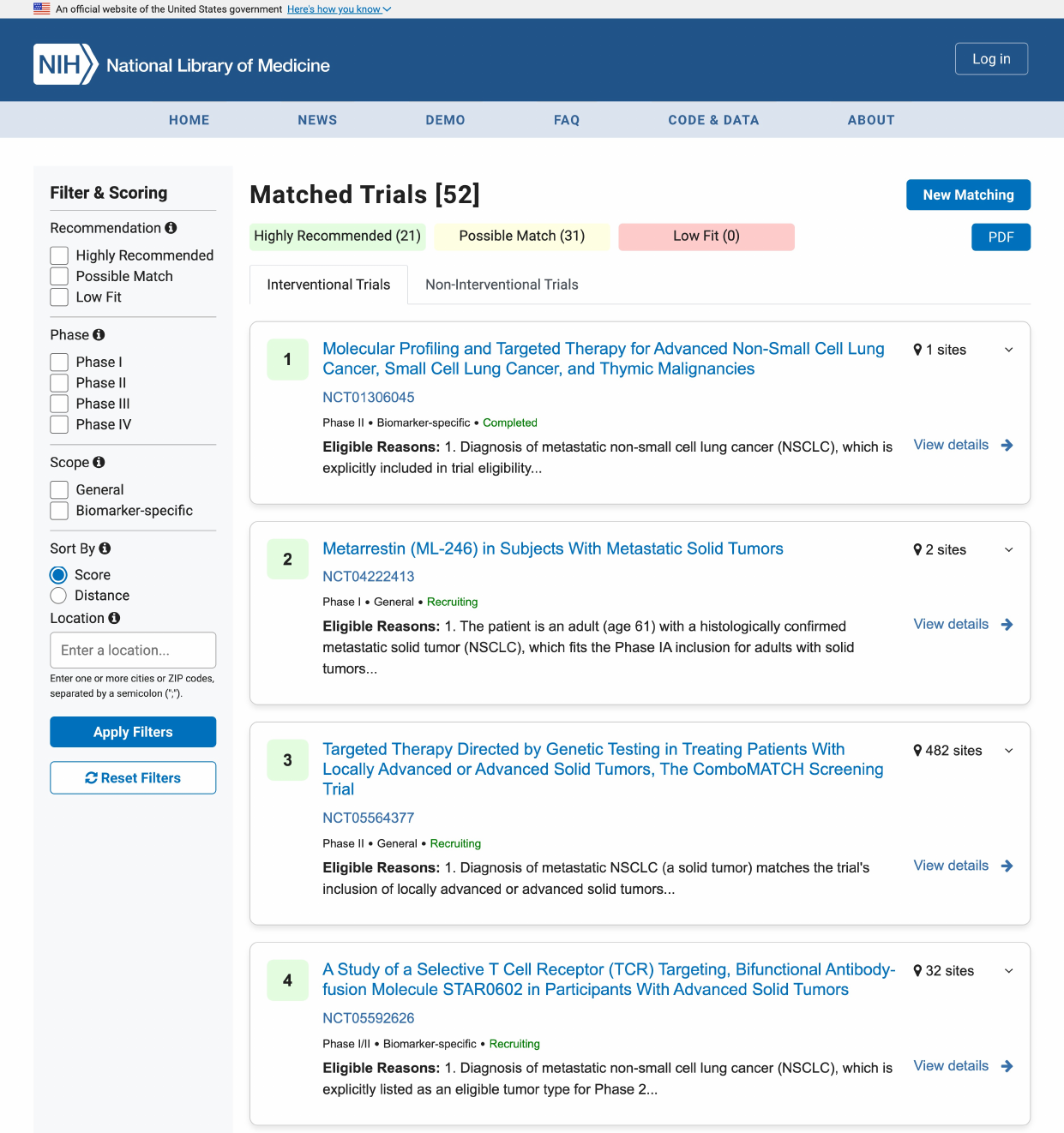}
  \caption{\textbf{TrialGPT 2.0 web-based interface for trial recommendation review.}
  The interface accepts a patient clinical summary and returns matched trials grouped by TrialGPT 2.0 recommendation category. Reviewers can filter and sort recommendations, export matched results and inspect trial-level explanations. Each trial card presents trial metadata and a structured rationale, including eligibility-supporting evidence, missing information and ineligibility concerns.
  }
  \label{fig:interface}
\end{figure}

To assess the operational feasibility of TrialGPT 2.0 in the interface setting (Supplementary Fig.~\ref{fig:interface}), we measured recommendation-generation time across five implementation settings: CCF, NCI, OPR, UIC and UPMC. For each source, we sampled five patient cases and ran each case three times using the corresponding site-specific trial corpus and matching configuration. Runs were performed in the interface runtime environment using 128 worker processes.

\begin{figure}[!ht]
  \centering
  \includegraphics[width=0.6\textwidth]{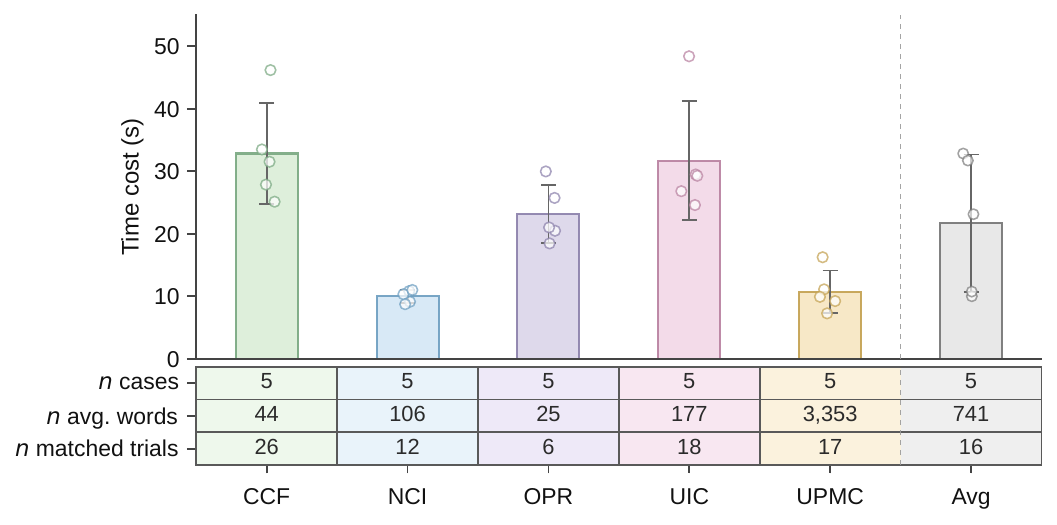}
  \caption{\textbf{TrialGPT 2.0 recommendation-generation runtime across evaluation sources.}
  Five cases are sampled from each evaluation source, and each case is run three times using 128 worker processes. The three runs are averaged to obtain a case-level runtime estimate. Bars show the mean runtime across cases within each source, error bars show the standard deviation across cases and points indicate case-level means; the Avg bar pools case-level means across all sources. Runtime includes trial retrieval when applicable, backend assessment, output parsing, result writing and generation of score-sorted PDF reports.
  }
  \label{fig:supplementary-time-cost}
\end{figure}

The timing experiment used the same trial-selection configuration as the interface. Retrieval was used only for UIC, where the trial corpus was first filtered using the hybrid retrieval strategy based on BM25 and MedCPT; by default, the top 267 retrieved trials were passed to the TrialGPT 2.0 backend for detailed assessment. For CCF, NCI, OPR and UPMC, retrieval was not used and TrialGPT 2.0 scored the full site-specific trial corpus directly. For UIC, retrieval was applied before backend assessment.

Timing was measured from initiation of the matching process to completion of the TrialGPT 2.0 recommendation-generation pipeline. The measurement included trial retrieval when applicable, backend assessment, output parsing, result writing and score-sorted PDF generation. Manual clinician review, downstream chart review and coordinator follow-up were not included.

For each case, we averaged the three repeated runs to obtain a case-level runtime estimate. We then summarized recommendation-generation time across cases within each source using the mean and standard deviation. Across all 25 sampled cases, TrialGPT 2.0 generated recommendations in 21.7 s on average, supporting use of the interface in trial-review workflows that require short turnaround time for candidate-trial generation (Supplementary Fig.~\ref{fig:supplementary-time-cost}).

\phantomsection\label{suppnote:reviewer-acceptance}
\subsection*{Supplementary Note 6: Clinical reviewer acceptance of TrialGPT 2.0}

\begin{table}[!ht]
\centering
\caption{\textbf{Clinical reviewer acceptance questionnaire.}}
\label{tab:clinical_reviewer_acceptance}
\small
\setlength{\tabcolsep}{3pt}
\renewcommand{\arraystretch}{1.35}

\textbf{Instructions:}
Please indicate your level of agreement with each statement based on your experience
using TrialGPT 2.0 for clinical trial review.

\vspace{0.4em}

\textbf{5-point Likert scale:}
1 = Strongly disagree, 2 = Disagree, 3 = Neutral, 4 = Agree, 5 = Strongly agree.

\vspace{0.8em}

\begingroup
\scriptsize
\arrayrulecolor{gray!55}
\setlength{\arrayrulewidth}{0.4pt}

\begin{tabularx}{0.98\linewidth}{|N|S|R|R|R|R|R|}
\hline
\rowcolor{gray!10}
\multicolumn{1}{|c|}{} &
\multicolumn{1}{c|}{} &
\multicolumn{5}{c|}{\textbf{Level of agreement (select one)}} \\
\cline{3-7}
\rowcolor{gray!10}
\multicolumn{1}{|c|}{\multirow{-2}{*}{\textbf{No.}}} &
\multicolumn{1}{c|}{\multirow{-2}{*}{\textbf{Statement}}} &
\makecell{\textbf{1}\\[-1pt]\scriptsize Strongly\\[-1pt]\scriptsize disagree} &
\makecell{\textbf{2}\\[-1pt]\scriptsize Disagree} &
\makecell{\textbf{3}\\[-1pt]\scriptsize Neutral} &
\makecell{\textbf{4}\\[-1pt]\scriptsize Agree} &
\makecell{\textbf{5}\\[-1pt]\scriptsize Strongly\\[-1pt]\scriptsize agree} \\
\hline
1 & \textbf{Usability:} TrialGPT 2.0 was user-friendly. & \choice & \choice & \choice & \choice & \choice \\
\hline
2 & \textbf{Workflow fit:} TrialGPT 2.0 fit well into my clinical workflow. & \choice & \choice & \choice & \choice & \choice \\
\hline
3 & \textbf{Efficiency:} TrialGPT 2.0 reduced the time required for trial review. & \choice & \choice & \choice & \choice & \choice \\
\hline
4 & \textbf{Clarity:} TrialGPT 2.0 outputs were easy to understand. & \choice & \choice & \choice & \choice & \choice \\
\hline
5 & \textbf{Decision confidence:} TrialGPT 2.0 increased my confidence in my trial-screening decisions. & \choice & \choice & \choice & \choice & \choice \\
\hline
6 & \textbf{Usefulness:} TrialGPT 2.0 outputs were relevant and meaningful for trial screening. & \choice & \choice & \choice & \choice & \choice \\
\hline
7 & \textbf{Content safety:} I did not observe violent, offensive, or otherwise harmful content in the TrialGPT 2.0 outputs I reviewed. & \choice & \choice & \choice & \choice & \choice \\
\hline
8 & \textbf{Satisfaction:} Overall, I was satisfied with TrialGPT 2.0. & \choice & \choice & \choice & \choice & \choice \\
\hline
\end{tabularx}
\endgroup
\end{table}

Clinical reviewer acceptance of TrialGPT 2.0 was assessed using an 8-item questionnaire after reviewers used or reviewed TrialGPT 2.0 outputs in clinical trial matching workflows. The questionnaire covered usability, workflow fit, perceived efficiency, clarity, decision confidence, usefulness, content safety and overall satisfaction. Each item used a 5-point Likert scale, where 1 indicated strongly disagree and 5 indicated strongly agree (Supplementary Table~\ref{tab:clinical_reviewer_acceptance}).

\begin{figure}[!t]
  \centering
  \includegraphics[width=\textwidth]{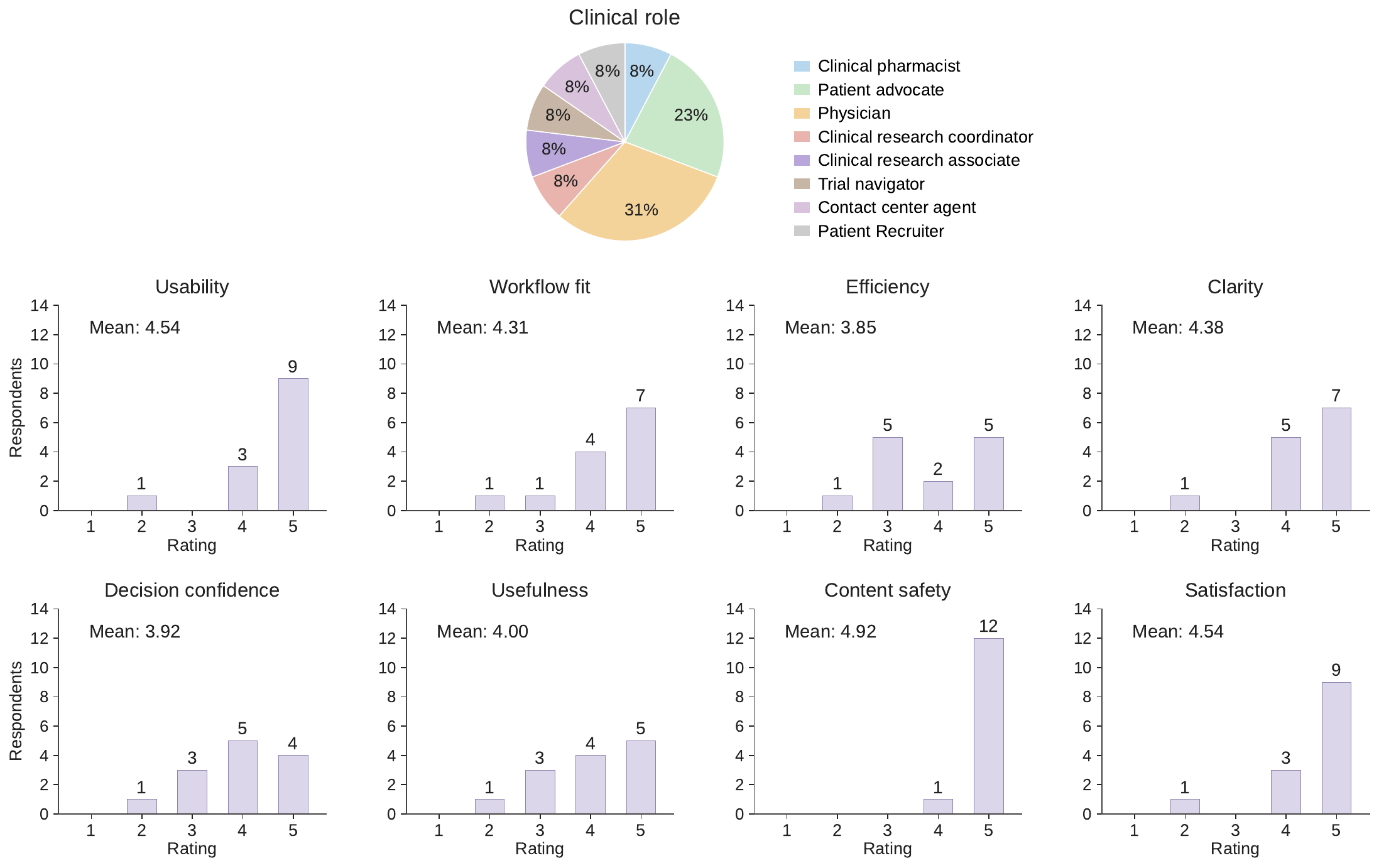}
  \caption{\textbf{Clinical reviewer acceptance responses for TrialGPT 2.0.}
  Clinical roles of questionnaire respondents and response distributions across the eight reviewer-acceptance items. The questionnaire was completed by 13 clinical reviewers after use or review of TrialGPT 2.0 outputs in clinical trial matching workflows. Items assessed usability, workflow fit, perceived efficiency, clarity, decision confidence, usefulness, content safety, and overall satisfaction on a 5-point Likert scale, where 1 indicates strongly disagree, and 5 indicates strongly agree. Bar plots show the number of respondents selecting each rating for each item; mean annotations indicate the average rating for each item.
  }
  \label{fig:supplementary-reviewer-acceptance-responses}
\end{figure}

\begin{figure}[h]
  \centering
  \includegraphics[width=0.8\textwidth]{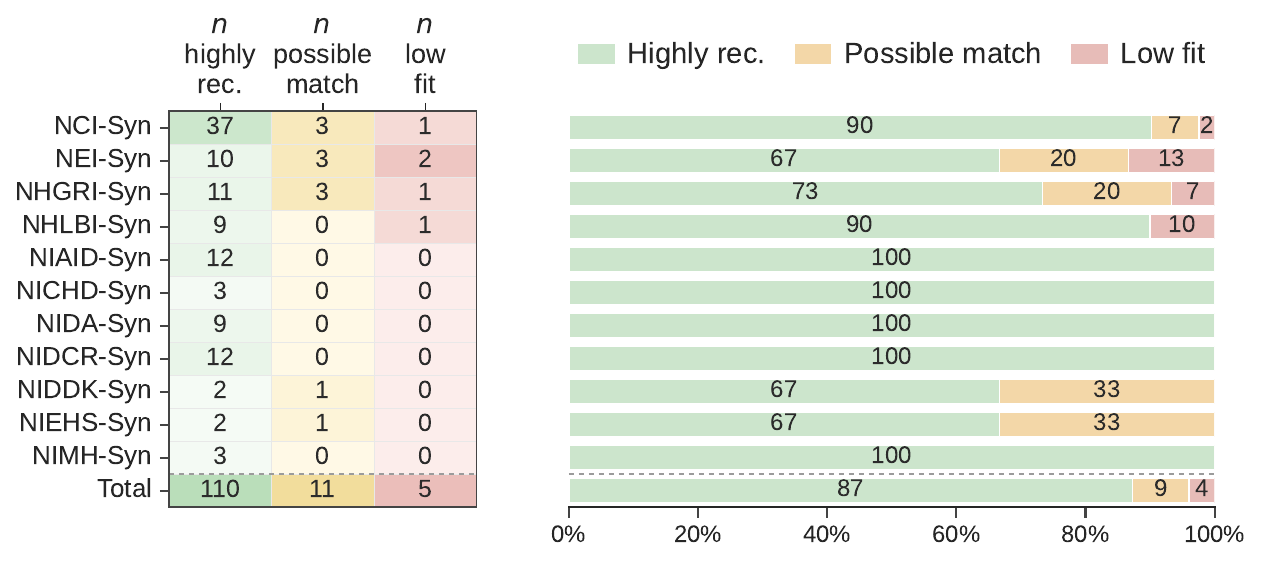}
  \caption{\textbf{NIH-TrialBench target-trial recovery.}
  Each synthetic vignette is associated with a prespecified target trial. For each vignette, we record the TrialGPT 2.0 recommendation category assigned to the target trial. The table and stacked bars show the number and proportion of target trials categorized as Highly Recommended, Possible Match, or Low Fit, separately for each contributing NIH Institute or Center and overall.
  }
  \label{fig:supplementary-synthetic-vignettes}
\end{figure}

Thirteen clinical reviewers completed the questionnaire. Across 104 item-level ratings, 85 ratings were favorable, defined as a score of 4 or 5. Mean item scores ranged from 3.85 to 4.92. Content safety received the highest rating, with a mean score of 4.92 and favorable responses from all reviewers. Overall satisfaction and usability each received a mean score of 4.54, followed by clarity and workflow fit, with mean scores of 4.38 and 4.31, respectively. Ratings were more variable for perceived efficiency and decision confidence, with mean scores of 3.85 and 3.92, respectively. These results suggest that reviewers generally found TrialGPT 2.0 safe, usable, understandable, and satisfactory, while perceived time savings and confidence gains may depend on workflow context, reviewer role, and continued user calibration (Supplementary Fig.~\ref{fig:supplementary-reviewer-acceptance-responses}).

\clearpage

\end{document}